\pdfoutput=1
\documentclass[10pt,twocolumn,letterpaper]{article}

\PassOptionsToPackage{table}{xcolor}

\usepackage[pagenumbers]{cvpr}   % To force page numbers, e.g. for an arXiv version
\usepackage{cuted}

\usepackage{currfile}
\usepackage{caption}
\usepackage{multirow}
\usepackage{array}
\usepackage{tabularx}
\usepackage{needspace}

\newlength{\maincol}
\newcommand{\mainhead}[1]{%
  \settowidth{\dimen0}{\scriptsize\textbf{#1}}%
  \ifdim\dimen0>\maincol
    \resizebox{\maincol}{!}{\scriptsize\textbf{#1}}%
  \else
    {\scriptsize\textbf{#1}}%
  \fi}

\usepackage{amssymb}
\usepackage{microtype}
\usepackage{placeins}
\usepackage{xspace}

\definecolor{headerbg}{RGB}{245,247,250}
\definecolor{oursbg}{RGB}{255,244,230}
\definecolor{pendingbg}{RGB}{250,250,250}

\newcommand{\method}{\textsc{SyncShare}\xspace}

\newcommand{\dpotwo}{DP$^2$O-FLUX}

\definecolor{cvprblue}{rgb}{0.21,0.49,0.74}
\usepackage[pagebackref,breaklinks,colorlinks,allcolors=cvprblue]{hyperref}

\def\confName{CVPR}
\def\confYear{2027}

\title{Every Model Has Its Moment: Multi-Diffusion Hypothesis Sharing for Real-World Image Super-Resolution}

\author{
Zhong Ji$^{1,*}$ \quad
Yiliang Zhao$^{2}$ \quad
Jingren Liu$^{1,3}$ \quad
Kanzhi Wu$^{3}$ \\
$^{1}$School of Electrical and Information Engineering, Tianjin University \\
$^{2}$International Engineering Institute, Tianjin University \quad $^{3}$vivo Mobile Ltd. \\
{\tt\small jizhong@tju.edu.cn}
}

\begin{document}
\maketitle

\begin{abstract}
Real-world image super-resolution (Real-ISR) systems commit to one diffusion restorer, though available restorers bring distinct priors and fail on different inputs. We inspect sampling trajectories: under a shared LQ-relative score, DiffBIR, SeeSR, DiT4SR, and \dpotwo{} trade the lead as denoising proceeds, changing leader at least once on $16$ of $20$ sampled RealSR\_512 images, so the best estimate is a moving target, not a fixed model. However, combining them so their strengths offset is difficult: their latent parameterizations, schedules, and state-transition rules preclude direct state exchange, so prior systems compare final outputs or chain restorers. Capturing it requires progress-aligned comparison and injection that leaves each recipient's native dynamics intact. \method{} meets both without training. Branches pause on a normalized-progress grid, ranking RGB hypotheses by relative gain over bicubic LR across four no-reference metrics. A loser receives only the displacement to the winner in its own encoder space, so agreement transfers nothing, and its native sampler resumes. Each branch keeps its own continuation beside at most one injected alternative, so the pool cannot converge onto one trajectory. On four real-world benchmarks, \method{} leads on CLIPIQA+, MUSIQ, MANIQA, and held-out TOPIQ-NR, lifting \dpotwo{} on RealLR200 from $0.7334/0.5824$ to $0.7738/0.6269$ in CLIPIQA+/MANIQA. Ablations place the gain in the protocol, not in pooling: dual-path relative injection outperforms final-stage sharing. Needing neither retraining nor a shared latent space, it extends to newly released restorers, turning a frozen pool into one stronger model.
\end{abstract}
\vspace{-2ex}
\section{Introduction}
Real-world image super-resolution (Real-ISR) recovers a high-resolution (HR) image from a low-resolution (LR) observation whose degradation is unknown and spatially varying. Blur, sensor noise, compression, sharpening, and color shifts often co-occur. Because the observation does not determine all lost detail, a restorer must preserve the structures the input supports while reconstructing what remains ambiguous. Diffusion models have become the mainstream tool for this reconstruction, as rapid progress in generative modeling supplies the detail it demands. Yet each system still commits to a single diffusion restorer, even though available restorers bring distinct priors and fail on different inputs.

\begin{figure}[t]
    \centering
    \includegraphics[width=0.95\columnwidth]{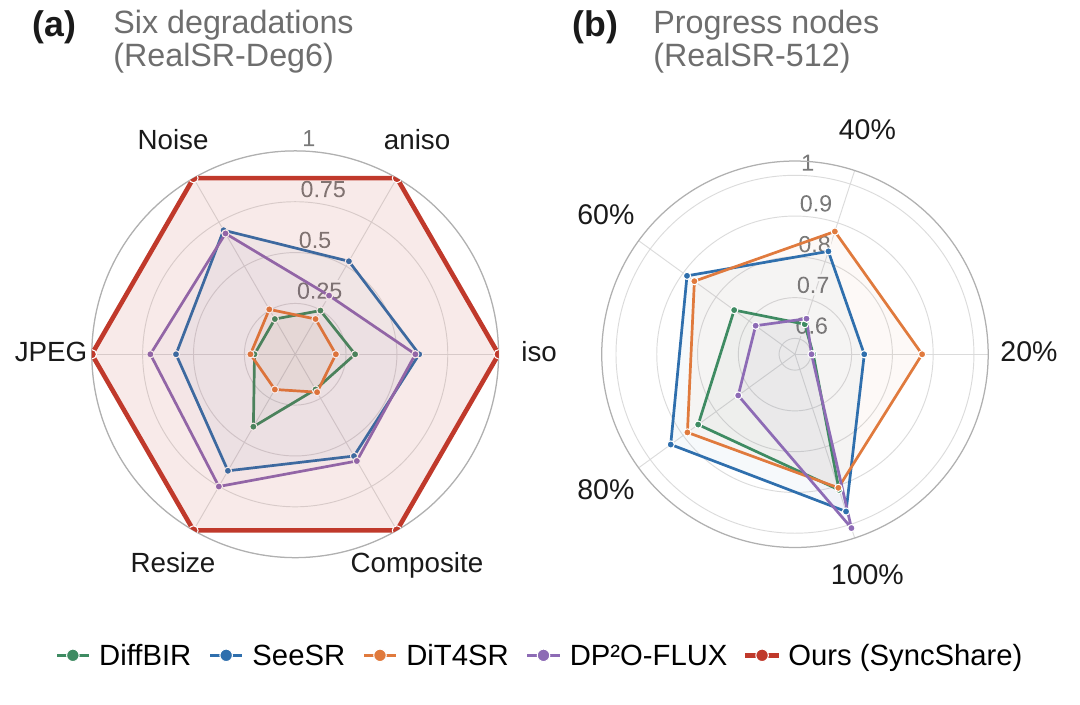}
    % \caption{Independent denoising on RealLR200 \#039. Each curve is that model's own RGB hypothesis under the LQ-relative score, and nothing is shared. DiT4SR leads through the middle, SeeSR overtakes it, and \dpotwo{} leads only at the finished output.}
    \caption{Motivation. (a) LQ-relative scores across six synthetic degradations added to RealSR\_512 (RealSR-Deg6), with \method{} in red for reference; no single restorer leads on every axis. (b) Independent trajectory scores across five progress nodes on RealSR\_512 images; the leader changes as denoising proceeds. Both studies use 20 uniformly sampled RealSR\_512 images.}
    \vspace{-3ex}
    \label{fig:motivation}
\end{figure}

% Their differences are not incidental. DiffBIR~\cite{lin2024diffbir} separates degradation removal from information regeneration; SeeSR~\cite{wu2024seesr} introduces degradation-aware semantic prompting; DiT4SR~\cite{duan2025dit4sr} adapts a diffusion transformer to Real-ISR; and DP$^2$O-SR~\cite{wu2025dp2osr} aligns generative restorers with perceptual preferences across diffusion- and flow-based backbones, with \dpotwo{} denoting its FLUX instantiation as named in the original work. Each contributes a different prior, and none dominates on every degradation (as shown in Figure~\ref{fig:motivation} (a)). What prevents them from being combined is mechanics rather than merit: they differ in architecture, latent parameterization, conditioning, sampling, text guidance, and in how LR information enters the trajectory, so their states cannot be exchanged directly. Prior multi-model systems therefore avoid exchanging states. Agent-based pipelines such as AgenticIR~\cite{zhu2025agenticir} and 4KAgent~\cite{zuo20254kagent} chain restoration tools in series, so each stage consumes the previous output: an early error cannot be revised, and both artifacts and latency accumulate downstream. Others compare finished images from independently run restorers, which ranks models without letting one exploit what another finds along the way. Neither design consults an intermediate estimate while its owner is still sampling. \method{} is built for that gap: it advances the restorers as parallel branches and lets them exchange hypotheses as sampling proceeds.
Their differences are not incidental. DiffBIR~\cite{lin2024diffbir} separates degradation removal from information regeneration; SeeSR~\cite{wu2024seesr} introduces degradation-aware semantic prompting; DiT4SR~\cite{duan2025dit4sr} adapts a diffusion transformer to Real-ISR; and DP$^2$O-SR~\cite{wu2025dp2osr} aligns generative restorers with perceptual preferences across diffusion- and flow-based backbones, with \dpotwo{} denoting its FLUX instantiation as named in the original work. Each contributes a different prior, and none dominates on every degradation (as shown in Figure~\ref{fig:motivation} (a)). What prevents them from being combined is mechanics rather than merit. They differ in architecture, latent parameterization, conditioning, sampling, text guidance, and in how LR information enters the trajectory, so their states cannot be exchanged directly. Prior multi-model systems therefore avoid exchanging states. Agent-based pipelines such as AgenticIR~\cite{zhu2025agenticir} and 4KAgent~\cite{zuo20254kagent} chain restoration tools in series, so each stage consumes the previous output; an early error cannot be revised, and both artifacts and latency accumulate downstream. 4KAgent also compares finished images from independently run restorers, which ranks models without letting one exploit what another finds along the way. Neither design consults an intermediate estimate while its owner is still sampling. \method{} is built for that gap. It advances the restorers as parallel branches and lets them exchange hypotheses as sampling proceeds.

% To measure what that choice discards, we score the four models' own hypotheses as they denoise one RealLR200 image, with nothing shared between them. Figure~\ref{fig:motivation} (b) traces the result: the LQ-relative scores rise in alternation, and the model in front changes as sampling proceeds. This pattern is not specific to one image. Over RealLR200 the leading hypothesis changes at least once on 197 of 200 images. There is therefore no single model to select in advance, since the strongest estimate migrates between models, while final-output comparison observes only where each model lands, never the stage at which it was worth borrowing from.
% To measure what that choice discards, Figure~\ref{fig:motivation} (b) scores the four models' unshared hypotheses on RealSR\_512 images: the LQ-relative scores alternate, and the leader shifts as sampling proceeds. This is not an isolated case—across the 20 sampled images, the leading hypothesis changes at least once on 16 of them. There is therefore no single model to select in advance, since the strongest estimate migrates between models, while final-output comparison observes only where each model lands, never the stage at which it was worth borrowing from.
To measure what that choice discards, Figure~\ref{fig:motivation} (b) scores the four models' unshared hypotheses on RealSR\_512 images, with each score relative to the low-quality input: the LQ-relative scores alternate, and the leader, that is the model holding the best score at a given stage, shifts as sampling proceeds. This is not an isolated case; across the 20 sampled images, the leading hypothesis changes at least once on 16 of them. No single model can therefore be chosen in advance, because the strongest estimate migrates between models along the trajectory, whereas final-output comparison observes only where each lands, never the stage at which borrowing from another model would have paid off.

The strongest estimate therefore changes owner as sampling proceeds, and a collaboration scheme has to borrow it before sampling ends. Three difficulties stand in the way. First, the models occupy incompatible state spaces: a latent from the UNet-style diffusers cannot be copied into the flow-matching transformers, and even VAE-based models differ in scaling factors, schedulers, and spatial conventions. Second, they use different denoising steps, so matching raw step indices does not align equivalent restoration stages. Third, arbitration needs a quality score, yet real-world data provide no HR target, and no-reference scores differ in scale and direction and remain sensitive to image content. Each difficulty has a specific answer, which \method{} provides.

\method is a training-free controller around the frozen restorers, built from three modules. A shared pause--share--resume protocol lets each branch export a resumable state and a clean-image hypothesis at chosen points in its own trajectory. \textbf{RelShift} handles incompatible state spaces by never moving a latent between models: each loser encodes the winner and its own hypothesis with its native encoder and receives only their \emph{relative latent displacement}, which is an exact no-op when the two agree, and the recipient's sampler turns that displacement into its next state in its native terms. \textbf{SyncGrid} handles unequal step counts by firing events at equal fractions of each model's own denoising budget rather than at equal step indices, so branches compare hypotheses only at equivalent restoration stages; its dual-path state machine also keeps a loser model's path alongside at most one injected path, bounding the pool at two branches per model. \textbf{AnchorRank} handles missing HR targets by ranking candidates with their relative gain over the resolution-matched bicubic-upsampled LR input across four no-reference metrics, placing metrics of different ranges and directions on one scale without any dataset-wide statistic. Together, these modules exploit complementary restoration priors without requiring the models to share a latent space or sampling schedule.

Our contributions are threefold:
\begin{itemize}
    \item A training-free pause--share--resume controller that enables heterogeneous diffusion restorers to exchange intermediate RGB hypotheses while preserving their pretrained, model-specific dynamics. To our knowledge, this is the first cross-model interaction of its kind during sampling.
    \item Three mechanisms make it usable: RelShift, transferring a recipient-space displacement to minimize cross-architecture conversion loss; SyncGrid, aligning unequal step counts by progress fraction and bounding the pool to two branches per model; and AnchorRank, ranking candidates by LQ-relative gains without dataset-wide statistics.
    \item A bounded dual-path hypothesis continuation machine that caps each model at two active branches. Together, the four components lead on CLIPIQA+, MUSIQ, and MANIQA across RealLR200, RealLQ250, RealSR\_512, and VideoLQ\_512, pointing to a concrete direction for agent-style interaction among restoration models.
    % A bounded dual-path design with at most two branches per model. Together, the four components lead on CLIPIQA+, MUSIQ, and MANIQA across RealLR\_200, RealLQ\_250, RealSR\_512, and VideoLQ\_512, pointing to a concrete direction for agent-style interaction among restoration models.
\end{itemize}
\section{Related Work}
\subsection{Real-World Image Super-Resolution}
Real-world image super-resolution (Real-ISR) extends super-resolution from fixed, known downsampling to photographs with unknown, spatially varying degradations. Early CNN and transformer restorers, including SRCNN~\cite{dong2016srcnn}, EDSR~\cite{lim2017edsr}, RCAN~\cite{zhang2018rcan}, SwinIR~\cite{liang2021swinir}, and HAT~\cite{chen2023hat}, typically assume a known downsampling. To move beyond this assumption, RealSR~\cite{cai2019realsr} constructs training pairs from real captures, while BSRGAN~\cite{zhang2021bsrgan} and Real-ESRGAN~\cite{wang2021realesrgan} approximate real degradations through higher-order synthesis. In parallel, SRGAN~\cite{ledig2017srgan} and ESRGAN~\cite{wang2018esrgan} optimize perceptual realism rather than distortion alone, but restoration remains governed by the perception--distortion tradeoff~\cite{blau2018perception}.

Diffusion-based restorers~\cite{saharia2022sr3,kawar2022ddrm,yue2023resshift,wang2024stablesr} address the perceptual side of this tradeoff by supplementing reconstruction with generative priors~\cite{ho2020ddpm,nichol2021improved,song2021ddim,song2021score,kingma2014vae,rombach2022ldm,podell2024sdxl,flux2024,peebles2023dit,lipman2023flow,liu2023rectified,esser2024sd3}.
Recent Real-ISR methods improve different aspects of the restoration process. DiffBIR~\cite{lin2024diffbir} separates degradation removal from information regeneration, SUPIR~\cite{yu2024supir} introduces ControlNet-style conditioning on the degraded image~\cite{zhang2023controlnet}, and DreamClear~\cite{ai2024dreamclear} focuses on constructing paired training data. Other methods modify the guidance, backbone, sampling budget, or training objective: SeeSR~\cite{wu2024seesr} adds degradation-aware semantic prompts, DiT4SR~\cite{duan2025dit4sr} adopts a diffusion transformer~\cite{peebles2023dit}, OSEDiff~\cite{wu2024osediff} reduces restoration to a single step, and DP$^2$O-SR~\cite{wu2025dp2osr} aligns the full trajectory with perceptual preferences. These design choices produce complementary restoration behavior, but they also create heterogeneous latent parameterizations, schedulers, and state-transition rules. The diversity that makes multiple restorers worth combining therefore prevents them from directly exchanging intermediate states.

The agent-based systems closest to our work coordinate restoration tools through their outputs. AgenticIR~\cite{zhu2025agenticir} uses a vision-language model to perceive the input and schedule restoration tools, while 4KAgent~\cite{zuo20254kagent} separates planning and tool execution for any-image-to-4K restoration. In these designs, each tool completes its current operation before its output is passed to a later stage or reviewed by the controller, rather than sharing intermediate estimates between ongoing diffusion trajectories. \method instead coordinates frozen restorers during sampling: decoded hypotheses are compared at matched normalized progress, and shared information is converted into each recipient's native representation before its sampler resumes.

\subsection{No-Reference Image Quality Assessment}
Intermediate arbitration requires no-reference image quality assessment because aligned high-quality references are unavailable during inference. Existing metrics capture different quality signals. NIQE~\cite{mittal2013niqe} and BRISQUE~\cite{mittal2012brisque} measure departures from natural-scene statistics, whereas NIMA~\cite{talebi2018nima}, MUSIQ~\cite{ke2021musiq}, MANIQA~\cite{yang2022maniqa}, and TOPIQ~\cite{chen2024topiq} learn quality predictions from human opinion scores. CLIPIQA+~\cite{wang2023clipiqa} and Q-Align~\cite{wu2024qalign} exploit vision-language models~\cite{radford2021clip}, and ImageReward~\cite{xu2023imagereward} learns preferences from human comparisons.
This diversity provides complementary assessments but complicates candidate comparison. Metrics differ in score range, preferred direction, and content sensitivity, making raw scores unsuitable for direct averaging and potentially biasing arbitration toward the aspect measured by a single metric. \method therefore ranks candidates by their average relative improvement over a common bicubic baseline across CLIPIQA+, NIQE, MUSIQ, and MANIQA, without requiring dataset-level normalization statistics. Q-Align and TOPIQ-NR, the no-reference variant of TOPIQ, are excluded from both arbitration and final reranking and used only to evaluate the selected outputs.
\section{Method}
\label{sec:method}

% \subsection{Overview}
% \label{sec:overview}
% Given a real-world LR input $I_{\mathrm{LQ}}$, our goal is to coordinate a pool of frozen pretrained diffusion restorers without mapping them into a common latent space. Let $\mathcal{M}=\{M_1,\dots,M_K\}$ denote the available models, all configured for the same upscale factor $r$. Each model $M_k$ has $T_k$ native inference steps and keeps its own state representation, scheduler, encoder, decoder, and conditioning mechanism, so its intermediate states occupy a model-specific domain $\mathcal{Z}_k$ that is neither shared with nor aligned to $\mathcal{Z}_j$ for $j\neq k$.
% Given a real-world LR input $I_{\mathrm{LQ}}$, our goal is to coordinate a pool of frozen pretrained diffusion restorers without mapping them into a common latent space. Let $\mathcal{M}=\{M_1,\dots,M_K\}$ denote the available models, all configured for the same upscale factor $r$. Each model $M_k$ has $T_k$ native inference steps and maintains its own state representation, scheduler, encoder, decoder, and conditioning mechanism, so its intermediate states occupy a model-specific domain $\mathcal{Z}_k$ that is neither shared with nor aligned to $\mathcal{Z}_j$ for $j\neq k$; coordination therefore cannot rely on a shared intermediate representation.
Given a real-world LR input $I_{\mathrm{LQ}}$, our goal is to coordinate a pool of frozen pretrained diffusion restorers without mapping them into a common latent space. Let $\mathcal{M}=\{M_1,\dots,M_K\}$ denote the available models, all configured for the same upscale factor $r$. Each model $M_k$ has $T_k$ native inference steps and keeps its own state representation, scheduler, encoder, decoder, and conditioning mechanism, so its intermediate states occupy a model-specific domain $\mathcal{Z}_k$ that is neither shared with nor aligned to $\mathcal{Z}_j$ for $j\neq k$. Coordination among the models can thus be mediated only by quantities that remain observable outside these private domains, since no intermediate state of one model can be transferred, fused, or directly compared against that of another.

% A model-specific adapter initializes a restoration session, advances it to the next pause, decodes its clean-state estimate, and builds a valid continuation from a shared RGB hypothesis, with the resulting state updates derived in Section~\ref{sec:method_reinjection}. Let $x_t^{(k)}\in\mathcal{Z}_k$ be the state entering step $t$, and write $\epsilon_t^{(k)}=\epsilon_\theta^{(k)}(x_t^{(k)},t)$ and $v_t^{(k)}=v_\theta^{(k)}(x_t^{(k)},\sigma_t)$ for the noise and velocity predictions~\cite{ho2020ddpm,lipman2023flow}. Both parameterizations expose a one-step estimate of the clean latent

A model-specific adapter initializes a restoration session, advances it to the next pause, decodes its clean-state estimate, and builds a valid continuation from a shared RGB hypothesis, with state updates derived in Section~\ref{sec:method_reinjection}. Let $x_t^{(k)}\in\mathcal{Z}_k$ be the state entering step $t$, and let $\epsilon_t^{(k)}=\epsilon_\theta^{(k)}(x_t^{(k)},t)$ and $v_t^{(k)}=v_\theta^{(k)}(x_t^{(k)},\sigma_t)$ denote the noise and velocity predictions~\cite{ho2020ddpm,lipman2023flow}. Both parameterizations yield a one-step estimate of the clean latent:
\begin{equation}
\begin{aligned}
    \hat{x}_0^{(k)}
    &=\mathcal{G}_k\!\left(x_t^{(k)};t\right)\\
    &=
    \begin{cases}
    \dfrac{x_t^{(k)}-\sqrt{1-\bar{\alpha}_t}\,\epsilon_t^{(k)}}
    {\sqrt{\bar{\alpha}_t}}, & \text{DDPM},\\[2mm]
    x_t^{(k)}-\sigma_t v_t^{(k)}, & \text{Flow Matching}.
    \end{cases}
\end{aligned}
\label{eq:clean_estimator}
\end{equation}

At normalized progress $\rho_k=n_k/T_k$ a branch has completed $n_k=\lfloor\rho_kT_k\rfloor$ native steps and reached timestep $t(\rho_k)$, so its hypothesis is $H_k(\rho_k)=D_k\big(\hat{x}_0^{(k)}(\rho_k)\big)$ with $\hat{x}_0^{(k)}(\rho_k)=\mathcal{G}_k\big(x_{t(\rho_k)}^{(k)};t(\rho_k)\big)$ and $D_k$ the native decoder~\cite{rombach2022ldm}. We call $H_k(\rho_k)$ a \emph{decoded hypothesis}: it estimates the final restored image while sampling is still incomplete. The latent domains are incompatible, yet the decoders share a common codomain, so hypotheses can be compared across models without any shared latent variable.

The quantity we exploit is the ordering induced by the arbitration score $S$ of Section~\ref{sec:selector}. Since the leader $k^{\star}(\rho)=\arg\max_k S\big(H_k(\rho)\big)$ generally varies with $\rho$, collaboration must be scheduled throughout sampling instead of being resolved at the final output, as Figure~\ref{fig:motivation} illustrates.

\begin{figure*}[t]
    \centering
    \includegraphics[width=0.91\textwidth]{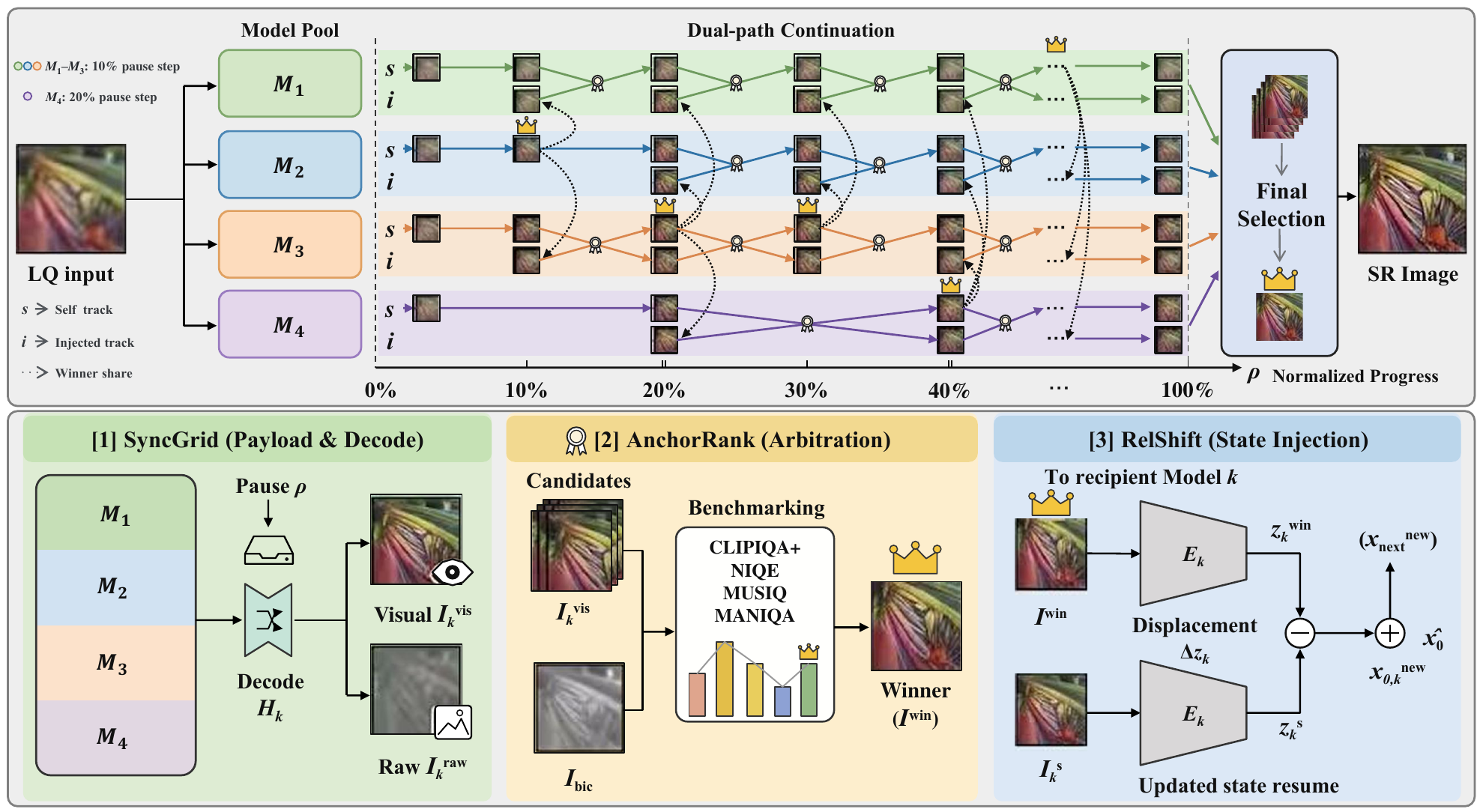}
    % \caption{Overview of \method. Frozen, heterogeneous restorers pause on a shared normalized-progress grid. The controller compares their decoded RGB hypotheses and transfers the winner to each losing model as a relative latent displacement in the recipient's native representation space.}
    \caption{Overview of \method. Frozen restorers pause on a shared progress grid, and the controller passes the best decoded hypothesis to each loser as a relative latent displacement in that model's own representation space.}
    \label{fig:framework}
\end{figure*}

The controller repeatedly advances, compares, shares, and resumes active branches until all of them terminate, keeping every weight frozen and every native state transition intact: SyncGrid (Section~\ref{sec:syncgrid}) fixes when branches communicate and which continuations survive, AnchorRank (Section~\ref{sec:selector}) selects the hypothesis to share, and RelShift (Section~\ref{sec:method_reinjection}) transfers it into each recipient. Figure~\ref{fig:framework} illustrates the cycle.

\subsection{SyncGrid}
\label{sec:syncgrid}
A collaboration event must preserve the state the native sampler needs rather than merely export an intermediate image. Suppressing the branch index for readability, a pause returns the payload $\mathcal{P}_k=(x_t,\hat{x}_0,x_{\mathrm{next}},t,i,T_k,\xi_k)$, where $x_t$ enters the current denoising step, $\hat{x}_0$ is the clean-state estimate, $x_{\mathrm{next}}$ is the propagated state from which sampling resumes, and the remaining entries are the timestep, native loop index, trajectory length, and model metadata $\xi_k$ covering valid spatial sizes and padding. The resumable state $\mathbf{s}_k=(i_{\mathrm{next}},x_{\mathrm{next}})$ is cached alongside.

Each pause exposes two forms of the hypothesis, $C_k=(I_k^{\mathrm{vis}},I_k^{\mathrm{raw}},\mathcal{P}_k,\mathbf{s}_k)$. The visualization image $I_k^{\mathrm{vis}}$ is cropped and aligned to the valid SR output for IQA scoring and qualitative inspection, while the raw image $I_k^{\mathrm{raw}}$ remains on the shared padded canvas used for reinjection. Keeping the two apart stops evaluation transforms from affecting model-to-model state transfer.

\subsubsection{Event Scheduling}
Raw loop indices do not identify equivalent restoration stages when models use different trajectory lengths, so we measure normalized progress $\rho_k=n_k/T_k$ from the completed native steps and always process the smallest pending value, $\rho^{\star}=\min_{C_k\in\mathcal{C}_{\mathrm{pending}}}\rho_k$. Only branches at $\rho^{\star}$ join the event and the others continue toward their next pause, so no comparison mixes restoration stages.

In our four-model pool, DiffBIR, SeeSR, and DiT4SR expose candidates every $10\%$ of their trajectories, while \dpotwo{} contributes every $20\%$, and the controller verifies that all pause intervals lie on the common grid. Figure~\ref{fig:sharing}(a) illustrates the resulting collaboration dynamics.

\subsubsection{Dual-Path Continuation}
Replacing every losing trajectory with the current winner would discard distinct model priors, so each model keeps at most two branches, $\mathcal{B}_k\subseteq\{\mathrm{s},\mathrm{i}\}$, where $\mathrm{s}$ denotes the \emph{self} (or \emph{retained incumbent}) branch that advances without reinjection at the current pause and $\mathrm{i}$ denotes the \emph{injected} branch built from the global winner. 
Every model starts with a single $\mathrm{s}$ branch; at each pause, intra-model arbitration selects the higher-scoring branch in $\mathcal{B}_k$ to become the new $\mathrm{s}$ (promoting $\mathrm{i}$ to $\mathrm{s}$ whenever $\mathrm{i}$ outscores $\mathrm{s}$) and discards the weaker one. 
The global winner is never reinjected into itself, whereas each global loser keeps its $\mathrm{s}$ branch untouched and spawns a fresh $\mathrm{i}$ branch by transferring the global winner relative to $\mathrm{s}$.
A model keeps its strongest continuation together with at most one winner-derived alternative, and the pool stays bounded by $2K$ branches without growing across events. Figure~\ref{fig:sharing}(b) shows local overtaking, promotion of an injected branch, and a later win by a retained continuation.

\subsection{AnchorRank}
\label{sec:selector}
AnchorRank selects the hypothesis to share. Because aligned HQ references are unavailable during inference, it combines four no-reference metrics, $\mathcal{Q}=\{\text{CLIPIQA+},\text{NIQE},\text{MUSIQ},\text{MANIQA}\}$. Their raw scores differ in range and preferred direction, so each candidate is compared against an input-specific reference, the PIL bicubic upsample $I_{\mathrm{bic}}$ of $I_{\mathrm{LQ}}$ at the valid $\times4$ SR size. All candidates and the baseline are scored at that size, and any resize needed for scoring uses the same bicubic filter without touching model inputs, shared states, or saved outputs.

For a higher-is-better metric $q$ the relative gain is:
\begin{equation}
    R_q(I)=\frac{m_q(I)-m_q(I_{\mathrm{bic}})}
    {\max\left(\left|m_q(I_{\mathrm{bic}})\right|,\epsilon\right)},
    \label{eq:relative_positive}
\end{equation}
where $\epsilon>0$ prevents division by zero; NIQE uses the reversed difference, since lower is better. 

The arbitration score is the equal-weight mean:
\begin{equation}
    S(I)=\frac{1}{4}\sum_{q\in\mathcal{Q}}R_q(I).
    \label{eq:selector}
\end{equation}

A positive $R_q$ indicates an improvement over the bicubic baseline under metric $q$. Normalizing by that single shared reference puts all four metrics on one scale without dataset-level statistics, and it keeps a candidate from being favored merely because its absolute score is high.

Candidates are first compared within each model, and only the stronger of its two branches enters the global comparison, from which Eq.~\eqref{eq:selector} selects the winner. Once sampling ends, every surviving branch is decoded and reranked by the same score, and the highest-scoring image is returned.

\subsection{RelShift}
\label{sec:method_reinjection}
RelShift transfers a winning hypothesis without copying an incompatible latent. Let $I_k^{\mathrm{s}}$ be the recipient's current best hypothesis, $I^{\mathrm{win}}$ the global winner, and $E_k$ the recipient's native encoder or packing operation. Encoding both, $z_k^{\mathrm{s}}=E_k(I_k^{\mathrm{s}})$ and $z_k^{\mathrm{win}}=E_k(I^{\mathrm{win}})$, brings them into the recipient's clean-state representation after model-specific scaling, packing, and valid-region conversion, and only their displacement is transferred:
\begin{equation}
    \Delta z_k=z_k^{\mathrm{win}}-z_k^{\mathrm{s}},
    \label{eq:general_relative}
\end{equation}
so the clean-state estimate moves to $x_{0,k}^{\mathrm{new}}=\hat{x}_{0,k}+\Delta z_k$ and the native scheduler builds the next resumable state from it.

This construction is an exact no-op when the two hypotheses agree, since $I^{\mathrm{win}}=I_k^{\mathrm{s}}$ gives $\Delta z_k=0$, and it cancels the shared component of a deterministic RGB-to-latent round trip instead of reading that representation offset as new information. The winner therefore shifts the recipient's own estimate, and the next state follows the recipient's sampler:
\begin{equation}
    x_{\mathrm{next},k}^{\mathrm{new}}=
    \begin{cases}
    \mu_t\big(x_t^{\mathrm{new}},x_0^{\mathrm{new}}\big)+\sigma_t\eta_t, & \text{DDPM},\\[1.5mm]
    x_{\mathrm{next}}+(1-\sigma_{\mathrm{next}})\,\Delta z_k, & \text{Flow Matching}.
    \end{cases}
    \label{eq:sampler_update}
\end{equation}

\subsubsection{DDPM Branches}
For DiffBIR and SeeSR, only the clean component of the transition changes. Following standard diffusion formulations~\cite{ho2020ddpm, song2021ddim}, recovering the noise implied by the original prediction, $\hat{\epsilon}_t=(x_t-\sqrt{\bar{\alpha}_t}\hat{x}_0)/\sqrt{1-\bar{\alpha}_t}$, rebuilds the state consistent with the shifted estimate, $x_t^{\mathrm{new}}=\sqrt{\bar{\alpha}_t}x_0^{\mathrm{new}}+\sqrt{1-\bar{\alpha}_t}\hat{\epsilon}_t$, where $x_0^{\mathrm{new}}=\hat{x}_0+\Delta z_k$. The next state is the posterior mean of $(x_t^{\mathrm{new}},x_0^{\mathrm{new}})$ plus a noise sample, and instead of drawing a new sample we recover the one the original transition already took, $\eta_t=\big(x_{\mathrm{next}}-\mu_t(x_t,\hat{x}_0)\big)/\sigma_t$. RelShift reuses the recipient's own randomness, and any change of trajectory is attributable to the injected displacement rather than to resampling.

\subsubsection{Flow-Matching Branches}
DiT4SR and \dpotwo{} use a FlowMatch Euler scheduler~\cite{lipman2023flow, liu2023rectified, esser2024sd3}. On the straight path $x_{\sigma}=(1-\sigma)x_0+\sigma\epsilon$ the network predicts $v=\epsilon-x_0$, which gives $\hat{x}_0=x_t-\sigma_tv_t$ and $x_{\mathrm{next}}=x_t+(\sigma_{\mathrm{next}}-\sigma_t)v_t$ at a pause. Since the same path reads $x_{\mathrm{next}}=(1-\sigma_{\mathrm{next}})x_0+\sigma_{\mathrm{next}}\epsilon$ at the next noise level, replacing $x_0$ by $\hat{x}_0+\Delta z_k$ at fixed $\epsilon$ shifts the next state by the coefficient that multiplies $x_0$, that is, by $(1-\sigma_{\mathrm{next}})\Delta z_k$. The winning hypothesis thus changes only the clean-state component of the next state, while the native flow-matching update governs the continuation.
\section{Experiments}
% \label{sec:experiments}
% \subsection{Experimental Setup}
\begin{table*}[!t]
\centering
\scriptsize
\setlength{\tabcolsep}{3.0pt}
\renewcommand{\arraystretch}{1.12}
\setlength{\aboverulesep}{0.2ex}\setlength{\belowrulesep}{0.3ex}
\setlength{\maincol}{33.0pt}%
\caption{No-reference results on four benchmarks. \method{} denotes the full four-model system. Methods are grouped into other single-model restorers, agent-based pipelines, and \method{}; the best value in each row is shown in red and the second best in blue underline. All metrics are higher-is-better except NIQE; Q-Align and TOPIQ-NR are held out from selection and reranking.}
\label{tab:main_results}
\begin{tabular}{@{}>{\scriptsize}w{c}{50pt}>{\scriptsize}w{c}{44pt}*{9}{>{\scriptsize}w{c}{\maincol}}>{\columncolor{oursbg}\scriptsize}w{c}{\maincol}@{}}
\toprule
\textbf{Dataset} & \textbf{Metric} & \mainhead{DiffBIR} & \mainhead{SeeSR} & \mainhead{DiT4SR} & \begin{tabular}[c]{@{}c@{}}\mainhead{DP$^2$O-}\\\mainhead{FLUX}\end{tabular} & \mainhead{CCSR} & \mainhead{VOSR} & \begin{tabular}[c]{@{}c@{}}\mainhead{Flux-}\\\mainhead{Upscaler}\end{tabular} & \mainhead{4KAgent} & \mainhead{AgenticIR} & \mainhead{SyncShare} \\
\midrule
\multirow{6}{*}{RealLR200} & CLIPIQA+ $\uparrow$ & 0.6960 & 0.6837 & 0.6948 & 0.7334 & 0.6963 & 0.5884 & 0.6693 & \textcolor{blue}{\underline{0.7396}} & 0.6744 & \textcolor{red}{\textbf{0.7738}} \\
& NIQE $\downarrow$ & 4.9759 & 4.1846 & \textcolor{blue}{\underline{3.8035}} & 4.0460 & 4.6719 & 4.3749 & 4.5270 & 3.8431 & \textcolor{red}{\textbf{3.7610}} & 4.4322 \\
& MUSIQ $\uparrow$ & 66.4319 & 69.5860 & 70.4418 & \textcolor{blue}{\underline{73.0939}} & 71.4697 & 68.1845 & 70.6343 & 71.4274 & 66.1685 & \textcolor{red}{\textbf{73.5202}} \\
& MANIQA $\uparrow$ & 0.5037 & 0.4977 & 0.4666 & \textcolor{blue}{\underline{0.5824}} & 0.4475 & 0.4284 & 0.4798 & 0.5341 & 0.4336 & \textcolor{red}{\textbf{0.6269}} \\
& Q-Align $\uparrow$ & 4.1197 & 4.2383 & 4.1958 & \textcolor{red}{\textbf{4.4423}} & 4.2668 & 4.0117 & 4.4119 & 4.2993 & 4.0946 & \textcolor{blue}{\underline{4.4344}} \\
& TOPIQ-NR $\uparrow$ & 0.6350 & 0.6632 & 0.6371 & \textcolor{blue}{\underline{0.7300}} & 0.6401 & 0.6025 & 0.6242 & 0.6867 & 0.5843 & \textcolor{red}{\textbf{0.7409}} \\
\midrule
\multirow{6}{*}{RealLQ250} & CLIPIQA+ $\uparrow$ & 0.7136 & 0.7063 & 0.7126 & \textcolor{blue}{\underline{0.7719}} & 0.6861 & 0.5944 & 0.6973 & 0.7490 & 0.6703 & \textcolor{red}{\textbf{0.7944}} \\
& NIQE $\downarrow$ & 5.1176 & 4.4532 & \textcolor{red}{\textbf{3.5388}} & 4.0989 & 4.9587 & 4.1833 & 4.4371 & 3.7349 & \textcolor{blue}{\underline{3.6435}} & 4.3105 \\
& MUSIQ $\uparrow$ & 67.5310 & 70.3760 & 71.8268 & \textcolor{blue}{\underline{73.8822}} & 72.0797 & 69.4886 & 71.0685 & 72.2769 & 64.8999 & \textcolor{red}{\textbf{74.8550}} \\
& MANIQA $\uparrow$ & 0.4900 & 0.4895 & 0.4597 & \textcolor{blue}{\underline{0.5863}} & 0.4255 & 0.4110 & 0.4826 & 0.5231 & 0.3906 & \textcolor{red}{\textbf{0.6150}} \\
& Q-Align $\uparrow$ & 3.9744 & 4.1413 & 4.2185 & 4.4027 & 4.0182 & 3.9128 & \textcolor{blue}{\underline{4.4448}} & 4.2705 & 4.0316 & \textcolor{red}{\textbf{4.4930}} \\
& TOPIQ-NR $\uparrow$ & 0.6355 & 0.6638 & 0.6421 & \textcolor{blue}{\underline{0.7351}} & 0.6413 & 0.5945 & 0.6289 & 0.6854 & 0.5573 & \textcolor{red}{\textbf{0.7460}} \\
\midrule
\multirow{6}{*}{RealSR\_512} & CLIPIQA+ $\uparrow$ & 0.6744 & 0.6818 & 0.6518 & 0.6702 & 0.6719 & 0.5860 & 0.5989 & \textcolor{blue}{\underline{0.7209}} & 0.6823 & \textcolor{red}{\textbf{0.7343}} \\
& NIQE $\downarrow$ & 5.7720 & 4.5597 & 4.3481 & 4.3586 & 5.1705 & 4.1082 & 5.1299 & \textcolor{blue}{\underline{4.0212}} & \textcolor{red}{\textbf{3.6353}} & 4.5301 \\
& MUSIQ $\uparrow$ & 65.3546 & 67.6006 & 67.7756 & 69.5665 & 69.2150 & 68.4609 & 67.5441 & \textcolor{blue}{\underline{70.4327}} & 66.2729 & \textcolor{red}{\textbf{71.8165}} \\
& MANIQA $\uparrow$ & 0.4992 & 0.5100 & 0.4499 & 0.5359 & 0.4663 & 0.4426 & 0.4489 & \textcolor{blue}{\underline{0.5440}} & 0.4496 & \textcolor{red}{\textbf{0.6000}} \\
& Q-Align $\uparrow$ & 3.9133 & 3.9955 & 3.9614 & 4.1101 & 4.1532 & 3.9511 & \textcolor{red}{\textbf{4.2484}} & 4.0834 & 4.0905 & \textcolor{blue}{\underline{4.1560}} \\
& TOPIQ-NR $\uparrow$ & 0.6168 & 0.6596 & 0.5888 & \textcolor{blue}{\underline{0.6858}} & 0.6285 & 0.6093 & 0.5731 & 0.6815 & 0.5907 & \textcolor{red}{\textbf{0.7190}} \\
\midrule
\multirow{6}{*}{VideoLQ\_512} & CLIPIQA+ $\uparrow$ & 0.5389 & 0.5791 & 0.5876 & 0.5628 & 0.6195 & 0.3666 & 0.4468 & \textcolor{blue}{\underline{0.6520}} & 0.5956 & \textcolor{red}{\textbf{0.6778}} \\
& NIQE $\downarrow$ & 4.8006 & 3.9224 & \textcolor{blue}{\underline{3.2552}} & 4.6881 & 4.2747 & 4.2921 & 4.5392 & 3.4237 & \textcolor{red}{\textbf{2.9769}} & 4.2269 \\
& MUSIQ $\uparrow$ & 58.1266 & 64.9766 & 66.7118 & 67.9619 & \textcolor{blue}{\underline{68.6979}} & 58.5717 & 62.3549 & 66.4093 & 62.5367 & \textcolor{red}{\textbf{69.9546}} \\
& MANIQA $\uparrow$ & 0.3985 & 0.4304 & 0.4086 & 0.4709 & 0.4070 & 0.3323 & 0.3638 & \textcolor{blue}{\underline{0.4827}} & 0.3952 & \textcolor{red}{\textbf{0.5469}} \\
& Q-Align $\uparrow$ & 3.8309 & 4.1072 & 4.1710 & \textcolor{blue}{\underline{4.2716}} & 4.1576 & 3.8523 & \textcolor{red}{\textbf{4.3361}} & 4.0582 & 4.2074 & 4.2613 \\
& TOPIQ-NR $\uparrow$ & 0.5669 & 0.6207 & 0.5966 & 0.6541 & 0.6152 & 0.5007 & 0.4997 & \textcolor{blue}{\underline{0.6559}} & 0.5628 & \textcolor{red}{\textbf{0.6882}} \\
\bottomrule
\end{tabular}
\vspace{-2ex}
\end{table*}
\noindent\textbf{Testing data.}
We evaluate real-world image enhancement on four benchmarks. RealLR200~\cite{wu2024seesr} contributes 200 in-the-wild LR images, while RealLQ250~\cite{ai2024dreamclear} provides 250 real low-quality images. RealSR\_512 comprises the 100 test captures from the $\times 4$ split of RealSR~\cite{cai2019realsr}, where each image is resized isotropically by $s=512/\max(W,H)$ if $\max(W,H)>512$ and left unchanged otherwise. VideoLQ\_512 is derived from the 50 VideoLQ clips of RealBasicVSR~\cite{chan2022realbasicvsr}. For a clip of $N$ frames we take the four temporally centered frames $j_i=\min\{N-1,\lfloor (i+0.5)N/4\rfloor\}$ with $i\in\{0,1,2,3\}$, giving 200 frames that are scored. Frames are EXIF-corrected, converted to RGB, and resized with Lanczos only when the long edge exceeds 512.

\noindent\textbf{Configurations.}
The collaborative pool contains four pretrained restorers: DiffBIR~\cite{lin2024diffbir}, SeeSR~\cite{wu2024seesr}, DiT4SR~\cite{duan2025dit4sr}, and \dpotwo~\cite{wu2025dp2osr}. We compare our system with these constituent models, three single-model references (CCSR~\cite{sun2025ccsr}, VOSR~\cite{wu2026vosr}, and Flux-Upscaler~\cite{jasperai2024fluxupscaler}), and two agent-based pipelines (4KAgent and AgenticIR). All runs use released weights without finetuning and produce $\times4$ outputs. DiffBIR and SeeSR use 50 steps with a pause every 5 steps, DiT4SR uses 40 steps with a pause every 4 steps, and \dpotwo{} uses 25 steps with a pause every 5 steps, corresponding to $10\%$, $10\%$, $10\%$, and $20\%$ normalized progress.

\noindent\textbf{Metrics.}
Selection and final reranking use CLIPIQA+, NIQE, MUSIQ, and MANIQA through the relative score of Eq.~\eqref{eq:selector}. Q-Align~\cite{wu2024qalign} and TOPIQ-NR~\cite{chen2024topiq} are computed on the selected image but influence neither decision. CLIPIQA+, MUSIQ, MANIQA, Q-Align, and TOPIQ-NR are higher-is-better, while NIQE is lower-is-better. 

% and Tables~\ref{tab:main_results} and~\ref{tab:ablation} report raw scores of the selected final images.

% Tables~\ref{tab:main_results} and~\ref{tab:ablation} report raw metric values rather than the relative gains used by the controller.

\noindent\textbf{Implementation details.}
A shared canvas padded on the right and bottom keeps intermediate RGB transfers spatially compatible across models, and the valid $\times4$ region is cropped before IQA. SeeSR concatenates RAM/DAPE tags with its positive prompt, DiT4SR uses benchmark-specific prompts, and the FLUX branch retains its native ControlNet-style conditioning and flow-matching scheduler. All runs use one NVIDIA RTX PRO 6000 (96\,GB), where the four frozen restorers advance sequentially.

\subsection{Quantitative Results}

% Table~\ref{tab:main_results} compares the full four-model system with single-model restorers and agent-based pipelines. All collaborative configurations use the recipient-space relative transfer in Section~\ref{sec:method_reinjection}; pairwise and triple-model pools are reported in Table~\ref{tab:pool} of the Appendix.
Table~\ref{tab:main_results} compares the full four-model \method{} system with single-model restorers and agent-based pipelines across four benchmarks. The single-model restorers include the four constituent models of \method{} and three additional standalone reference models. Results for the pairwise and triple-model pools are reported in Table~\ref{tab:pool} in the Appendix.

The full system achieves the highest CLIPIQA+, MUSIQ, and MANIQA on every benchmark. Against the strongest competitor for each benchmark and metric, CLIPIQA+ improves by $+0.0342$, $+0.0225$, $+0.0134$, and $+0.0258$, MANIQA by $+0.0445$, $+0.0287$, $+0.0560$, and $+0.0642$, and MUSIQ by $+0.4263$, $+0.9728$, $+1.3838$, and $+1.2567$ on RealLR200, RealLQ250, RealSR\_512, and VideoLQ\_512. The strongest competitor varies by benchmark and metric. For example, 4KAgent leads the competing methods in CLIPIQA+ and CCSR leads in MUSIQ on VideoLQ\_512.

NIQE favors other methods: AgenticIR leads on RealLR200 ($3.7610$), RealSR\_512 ($3.6353$), and VideoLQ\_512 ($2.9769$), while DiT4SR leads on RealLQ250 ($3.5388$). \method{} therefore does not dominate every metric. This outcome is consistent with its multi-metric objective, which balances four relative gains rather than optimizing NIQE alone.
The held-out metrics provide complementary checks. TOPIQ-NR ranks \method{} first on all four benchmarks, with margins of $+0.0109$, $+0.0109$, $+0.0332$, and $+0.0323$ over the respective runners-up. Q-Align produces a mixed ordering: \method{} leads on RealLQ250 ($4.4930$), trails \dpotwo{} by $0.0079$ on RealLR200, and trails Flux-Upscaler by $0.0924$ and $0.0748$ on RealSR\_512 and VideoLQ\_512. The improvements therefore do not transfer uniformly to every evaluator.
Table~\ref{tab:pool} in Appendix shows that two- and three-model pools already improve over their members. The full pool gives the highest MANIQA on all four benchmarks, although some triple pools retain slightly higher CLIPIQA+ on RealSR\_512 and VideoLQ\_512.

\begin{figure*}[!t]
    \centering
    \includegraphics[width=0.94\textwidth]{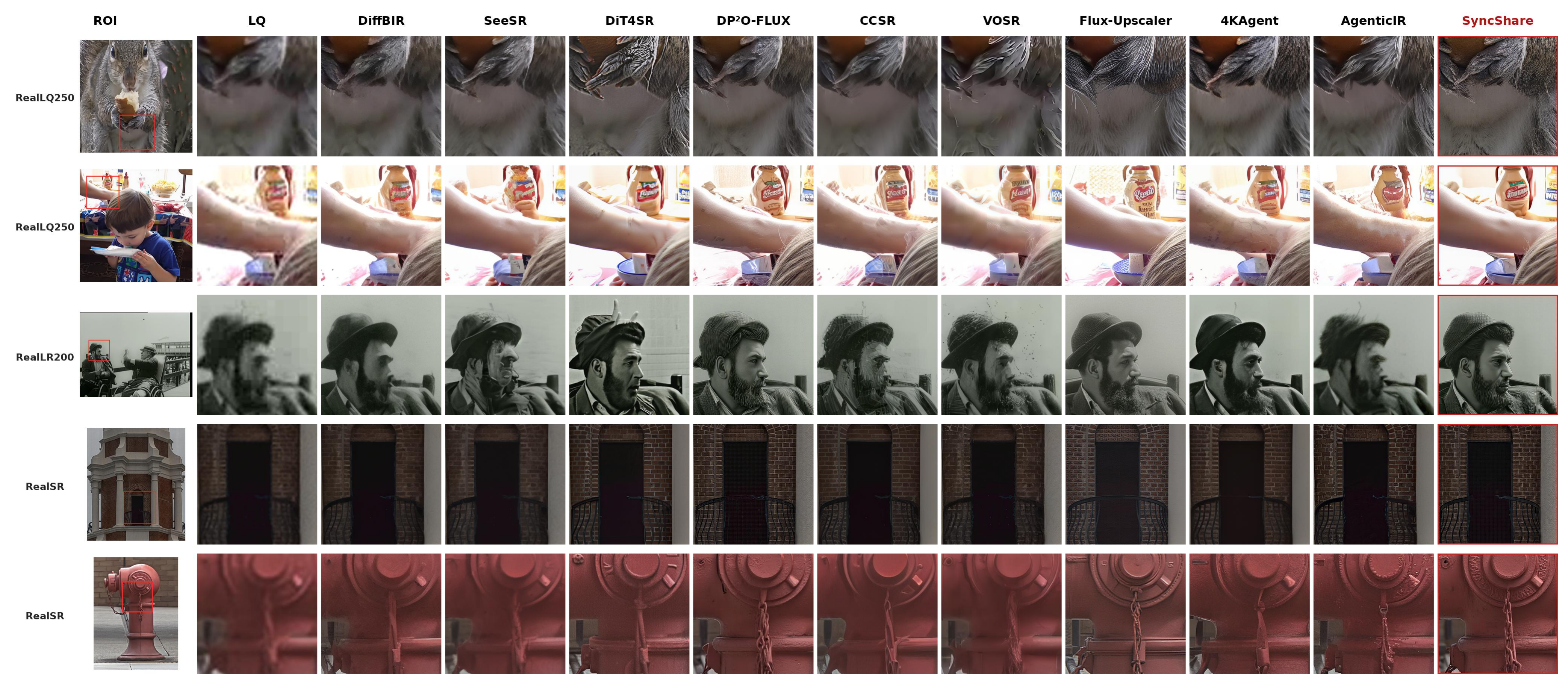}
    \caption{Qualitative comparison on still-image benchmarks (RealLQ250, RealLR200, RealSR\_512). Each row shows the full image with its red region of interest, then the LQ input and ten restorers on the same crop; \method{} is last. Best viewed at high magnification.}
    \label{fig:qualitative}
\end{figure*}

\begin{figure*}[!t]
    \centering
    \includegraphics[width=0.94\textwidth]{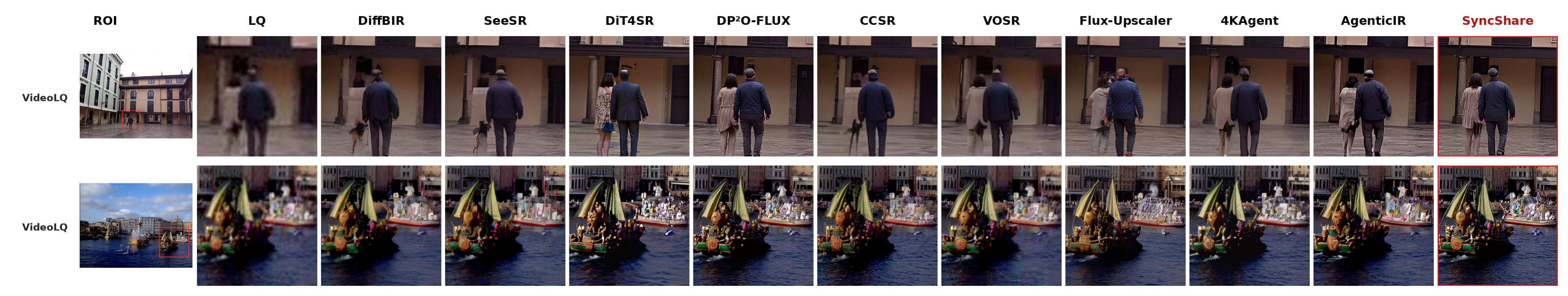}
    \caption{Per-frame comparison on VideoLQ\_512, without temporal fusion. Each row shows the full frame with its red region of interest, then the LQ input and ten restorers on the same crop; \method{} is last. Best viewed at high magnification.}
    \label{fig:qualitative_videolq}
\end{figure*}

\noindent\textbf{Per-image consistency.}
Aggregates can hide uneven behaviour, so we score each image against the strongest single restorer of its benchmark. \method{} wins on $78.5\%$, $64.4\%$, $74.0\%$, and $79.0\%$ of images respectively, with mean gains of $0.0635$, $0.0578$, $0.0805$, and $0.1289$ in the score of Eq.~\eqref{eq:selector} (Wilcoxon signed-rank $p<10^{-8}$ on all sets, paired Cohen's $d$ of $0.49$, $0.29$, $0.63$, and $0.70$). The deltas are right-skewed, with medians of $0.038$, $0.014$, $0.066$, and $0.098$, and a minority regress by up to $0.33$, so the mean is not carried by a few samples. Against a per-image oracle that may pick the best method for every image, \method{} still wins on $66.5\%$, $58.8\%$, $53.0\%$, and $68.0\%$. Per-metric win rates follow the aggregate picture, with CLIPIQA+ and MANIQA improving on $63$ to $83$ percent of images and NIQE on only $18$ to $46$ percent, the trade-off the selector makes by design. Figures~\ref{fig:delta} and~\ref{fig:winrate} of the Appendix give the distributions and the per-restorer win rates.

\noindent\textbf{Inference cost.}
Cost is measured on six images with the same GPU. A single diffusion restorer needs $11$ to $36$\,s per image, CCSR $0.7$\,s, AgenticIR $37$\,s, 4KAgent $224$\,s, and the four-model pool under \method{} $190$\,s with a peak of $84$\,GB. Training-free does not mean cheap. Keeping four models resident and paying decoding, scoring, and reinjection at every event puts our latency near the agent baseline and several times above a single restorer, so the method removes training rather than computation. Peak memory grows with the pool, from $58$\,GB for two models to $84$\,GB for four. Table~\ref{tab:cost} in Appendix lists every method.

\begin{table*}[htbp]
\centering
\scriptsize
\setlength{\tabcolsep}{3.6pt}
\renewcommand{\arraystretch}{1.12}
\caption{Controller-component ablation on RealLR200. Share: intermediate pause--share--resume; Dual: bounded two-branch ($\mathrm{s}$/$\mathrm{i}$) continuation; Rel.: recipient-space relative transfer instead of absolute injection; Rel-IQA: LQ-relative multi-metric arbitration instead of fixed-scale IQA. A dash marks a component that is undefined in that row. Columns without a dagger report the four-model pool, and columns with a dagger report the SeeSR+DiT4SR+DP$^2$O-FLUX pool. The best value is shown in red and the second best in blue underline.}
\label{tab:ablation}
\begin{tabular}{@{}>{\scriptsize}w{c}{64pt}cccc|cccc|cccc@{}}
\toprule
\rowcolor{headerbg}
\textbf{Configuration} & \textbf{Share} & \textbf{Dual} & \textbf{Rel.} & \textbf{Rel-IQA} &
\textbf{CLIPIQA+}$\uparrow$ & \textbf{NIQE}$\downarrow$ & \textbf{MUSIQ}$\uparrow$ & \textbf{MANIQA}$\uparrow$ &
\textbf{CLIPIQA+}$^{\dagger}\uparrow$ & \textbf{NIQE}$^{\dagger}\downarrow$ & \textbf{MUSIQ}$^{\dagger}\uparrow$ & \textbf{MANIQA}$^{\dagger}\uparrow$ \\
\midrule
Final-only & $\times$ & -- & -- & -- &
0.7570 & \textcolor{blue}{\underline{4.1513}} & \textcolor{red}{\textbf{73.7137}} & 0.6109 &
0.7567 & \textcolor{blue}{\underline{4.1609}} & \textcolor{red}{\textbf{73.7055}} & 0.6106 \\
Single-path+Abs & \checkmark & $\times$ & $\times$ & \checkmark &
0.7522 & 4.8420 & 71.6735 & 0.5860 &
0.7488 & 4.6225 & 71.4685 & 0.5765 \\
Dual+Abs & \checkmark & \checkmark & $\times$ & \checkmark &
0.7642 & 4.2850 & 73.5048 & \textcolor{blue}{\underline{0.6206}} &
\textcolor{blue}{\underline{0.7622}} & 4.2735 & 73.3954 & \textcolor{blue}{\underline{0.6191}} \\
\rowcolor{oursbg}
Full & \checkmark & \checkmark & \checkmark & \checkmark &
\textcolor{red}{\textbf{0.7738}} & 4.4322 & \textcolor{blue}{\underline{73.5202}} & \textcolor{red}{\textbf{0.6269}} &
\textcolor{red}{\textbf{0.7680}} & 4.3262 & \textcolor{blue}{\underline{73.5429}} & \textcolor{red}{\textbf{0.6234}} \\
Full+Fixed-IQA & \checkmark & \checkmark & \checkmark & $\times$ &
\textcolor{blue}{\underline{0.7649}} & \textcolor{red}{\textbf{3.9735}} & 73.4530 & 0.6099 &
0.7619 & \textcolor{red}{\textbf{3.9192}} & 73.3664 & 0.6060 \\
\bottomrule
\end{tabular}
\vspace{-3ex}
\end{table*}

\subsection{Qualitative Comparison}
Figure~\ref{fig:qualitative} compares shared crops from RealLQ250, RealLR200, and RealSR\_512. In these examples, DiffBIR, SeeSR, and CCSR leave some fine textures under-recovered, whereas DiT4SR can over-enhance or synthesize high-frequency structures unsupported by the input. \dpotwo{}, VOSR, and Flux-Upscaler exhibit different grain, sharpening, and structural artifacts. 
Among the agent pipelines, 4KAgent recovers sharper edges but still inherits the local blur or distortions of whichever finished tool output it selects, whereas AgenticIR produces smoother surfaces at the expense of fine details (such as fur strands, brickwork, and chain links), consistent with its lower NIQE in Table~\ref{tab:main_results}.
By exchanging hypotheses during sampling, \method{} preserves visible fur, facial features, building edges, and metal textures with less grain and fewer invented high-frequency structures. Figure~\ref{fig:qualitative_videolq} extends the comparison to VideoLQ\_512 frames without temporal fusion, where \method{} recovers clearer details on the person and boat crops, further highlighting its advantage in perceptual quality.

\subsection{Ablation Study}
\label{sec:ablation}

We isolate components on RealLR200 with all weights fixed. Table~\ref{tab:ablation} evaluates the four-model pool and a three-model pool of SeeSR, DiT4SR, and \dpotwo{}. \emph{Share} enables intermediate pause--share--resume; \emph{Dual} retains the self ($\mathrm{s}$) and injected ($\mathrm{i}$) continuations; \emph{Rel.} replaces absolute injection with recipient-space relative transfer; and \emph{Rel-IQA} replaces fixed-scale IQA with LQ-relative arbitration.

\noindent\textbf{Sharing versus final-output reranking.}
The first row reranks final outputs without intermediate exchange. The complete protocol consistently improves CLIPIQA+ over this baseline by $+0.0168$ and $+0.0113$, and MANIQA by $+0.0160$ and $+0.0128$, in the four- and three-model pools, respectively. This comparison evaluates the complete protocol rather than isolating Share alone.

\noindent\textbf{Dual-path continuation.}
With absolute injection and Rel-IQA fixed, adding Dual raises four-model CLIPIQA+ from $0.7522$ to $0.7642$, MUSIQ from $71.6735$ to $73.5048$, and MANIQA from $0.5860$ to $0.6206$, while reducing NIQE from $4.8420$ to $4.2850$. The three-model pool shows the same pattern. Retaining the stronger self branch ($\mathrm{s}$) alongside the injected path ($\mathrm{i}$) avoids the loss caused by overwriting the only active trajectory at every pause.

\noindent\textbf{Relative latent transfer.}
With Dual and Rel-IQA fixed, replacing absolute injection by relative transfer increases four-model CLIPIQA+/MANIQA from $0.7642/0.6206$ to $0.7738/0.6269$ and three-model CLIPIQA+/MANIQA from $0.7622/0.6191$ to $0.7680/0.6234$. MUSIQ also increases slightly in both pools, whereas NIQE becomes worse, revealing a metric-dependent trade-off.

\noindent\textbf{Relative IQA arbitration.}
The last row replaces the LQ-relative score with a \emph{fixed-scale} one: each raw metric is divided by a preset scale (1 for CLIPIQA+ and MANIQA, 100 for MUSIQ, and NIQE is mapped to $1-\mathrm{NIQE}/10$), averaging these higher-is-better terms equally without an LQ baseline. For the four-model pool, CLIPIQA+ falls from $0.7738$ to $0.7649$ and MANIQA from $0.6269$ to $0.6099$. Overall, the full configuration achieves the best CLIPIQA+ and MANIQA in both pools, while final-only reranking retains slightly higher MUSIQ.

\begin{table}[t]
\centering
\scriptsize
\setlength{\tabcolsep}{3.6pt}
\renewcommand{\arraystretch}{1.12}
\setlength{\aboverulesep}{0.2ex}\setlength{\belowrulesep}{0.3ex}
\setlength{\maincol}{33.0pt}%
\caption{Branch budget on the strided 10-image RealLR200 subset. $S$ is the
composite relative score of Eq.~\eqref{eq:selector} and $\Delta S$ is relative to
the dual-path default; the best value in each column is shown in red and the
second best in blue underline.}
\label{tab:budget}
\begin{tabular}{@{}>{\scriptsize}w{c}{64pt}*{4}{>{\scriptsize}w{c}{\maincol}}@{}}
\toprule
\rowcolor{headerbg}
\textbf{Budget} & \mainhead{$S$ (4-model)$\uparrow$} & \mainhead{$\Delta S$ (4-model)} & \mainhead{$S$ (3-model)$^{\dagger}\uparrow$} & \mainhead{$\Delta S$ (3-model)$^{\dagger}$} \\
\midrule
$B{=}1$ & 0.9799 & $-0.0042$ & 0.9504 & $-0.0339$ \\
\rowcolor{oursbg}
$B{=}2$ (default) & 0.9841 & 0.0000 & 0.9843 & 0.0000 \\
$B{=}3$ & \textcolor{red}{\textbf{0.9949}} & $+0.0108$ & \textcolor{blue}{\underline{0.9913}} & $+0.0070$ \\
\bottomrule
\end{tabular}
\vspace{-4ex}
\end{table}

\noindent\textbf{Schedule, budget, and selector controls.}
Table~\ref{tab:budget} varies the per-model branch budget on the uniformly strided 10-image RealLR200 subset (\texttt{\#001}, \texttt{\#021}, \dots, \texttt{\#181}), scoring each setting by $S$ of Eq.~\eqref{eq:selector} in the four- and three-model pools: the budget is a trade-off rather than a saturation point, since $B{=}1$ is nearly free with four models ($0.9799$) but costs $0.0339$ with three ($0.9504$), while $B{=}3$ adds $0.0108$ and $0.0070$. The other three control axes are reported in Table~\ref{tab:controls}. Coarsening the pause grid to $20\%$ costs $0.0043$ and $0.0048$, and $40\%$ costs a further $0.0031$ and $0.0023$, so the $10\%$ grid is kept; triggering pauses by raw step index agrees in direction but by a narrower margin ($0.0012$). The selector axis is decisive: NIQE alone falls to $0.9015$, while any single one of CLIPIQA+, MUSIQ, or MANIQA as the arbiter gives only $0.9495$--$0.9685$ with four models and $0.9431$--$0.9766$ with three, and dropping CLIPIQA+ from the four-metric set improves $S$ by at most $0.0131$. We retain all four metrics to preserve their complementary quality cues and provide a broader basis for visual assessment.
\FloatBarrier % Keep experiment floats (especially Table 2) before the Conclusion and References
\section{Conclusion}
% We present \method, a training-free controller that lets heterogeneous diffusion restorers interact during sampling rather than only at the output. To our knowledge, it is the first framework of its kind for Real-ISR: frozen models exchange decoded hypotheses and resume their own trajectories without shared latents or retraining. SyncGrid aligns events by progress fraction and bounds each model to two branches, AnchorRank arbitrates by LQ-relative gains, and RelShift transfers a hypothesis as a displacement in the recipient's encoder space. With four restorers, it leads on most metrics across four Real-ISR benchmarks.
% We present \method, a training-free controller that lets heterogeneous diffusion restorers interact during sampling rather than only at the output. \method is, to our knowledge, the first framework of its kind for Real-ISR: frozen models exchange decoded hypotheses and resume their own trajectories, without shared latents or retraining. The controller combines three mechanisms: SyncGrid aligns events by progress fraction and bounds each model to two branches; AnchorRank arbitrates by LQ-relative gains; and RelShift transfers a hypothesis as a displacement in the recipient's encoder space. With a pool of four restorers, \method leads on most metrics across four Real-ISR benchmarks.
We present \method, a training-free controller that lets heterogeneous diffusion restorers interact during sampling rather than only at the output. Such interaction lets a model revise its trajectory before committing to an output, a possibility that output-level combination forecloses. \method is, to our knowledge, the first framework of its kind for Real-ISR: frozen models exchange decoded hypotheses and resume their own trajectories, without shared latents or retraining. Off-the-shelf checkpoints are reused unmodified. The controller combines three mechanisms. SyncGrid aligns events by progress fraction, so that models with different step counts and schedulers agree on when to exchange, and bounds each model to two branches; AnchorRank arbitrates by LQ-relative gains, measuring each candidate's gain against the LQ input rather than in a latent space the models do not share; and RelShift transfers a hypothesis as a displacement in the recipient's encoder space, which requires no shared coordinate system.

\section{LLM Usage Statement}
\label{sec:6_LLMUsageStatement}
ChatGPT (OpenAI) was used to assist with revising manuscript text. Codex (OpenAI) was used to assist with research code debugging. The authors take full responsibility for the final manuscript, code, and reported results.

{
    \small
    \bibliographystyle{ieeenat_fullname}
    \bibliography{main}
}

% arXiv 版本建议保留补充材料，合并为一个完整的 PDF
\clearpage
\appendix
\raggedbottom
% \leftlinenumbers
% \nolinenumbers
\begin{strip}
  \begin{center}
    {\Large\textbf{\thetitle}}\\[0.6em]
    {\large Supplementary Material}
  \end{center}
  \vspace*{1.2em}
\end{strip}
% \linenumbers

% Float counters continue from the main paper (Figure A.5, Table B.4,
% Tables C.5--C.6, Figures C.6--C.7) to match the submitted main text.
\renewcommand{\thefigure}{\thesection.\arabic{figure}}
\renewcommand{\thetable}{\thesection.\arabic{table}}

\section{Sharing along the Trajectory}
\label{sec:app_sharing}
\begin{figure*}[t]
    \centering
    \includegraphics[width=0.9\textwidth]{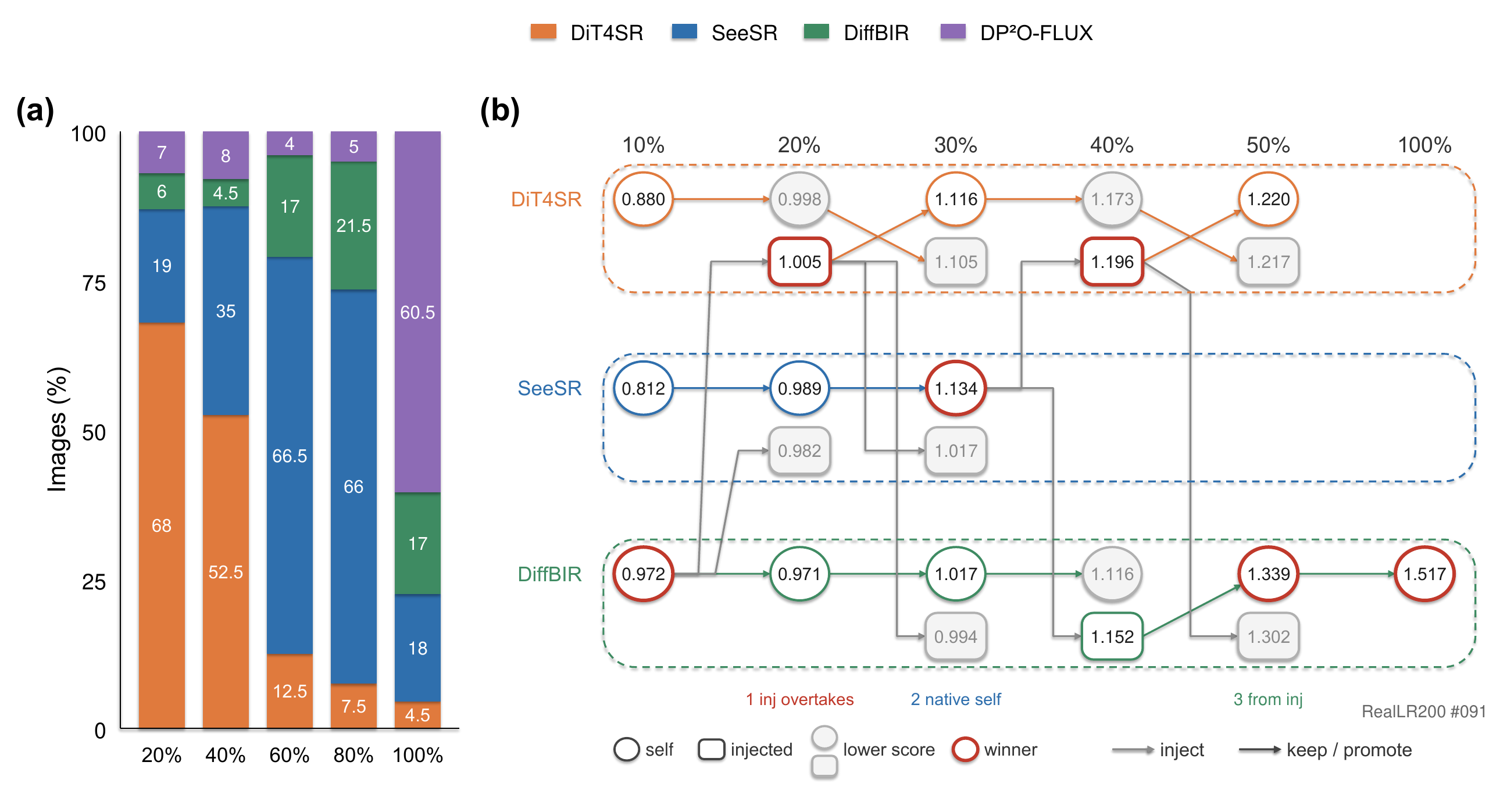}
    \vspace{-2ex}
    \caption{Sharing along the denoising trajectory on RealLR200. (a) Percentage of images won by each model at events where all four models participate (the 20\% grid) and at the 100\% final rerank. (b) Branch scores $S$ for image \#091. For legibility, \dpotwo{}, intermediate pauses after 50\%, and branches outside the three marked cases are omitted; labels 1--3 correspond to the cases in Section~\ref{sec:app_sharing}.}
    \label{fig:sharing}
\end{figure*}

Figure~\ref{fig:sharing} traces how the leading model changes during denoising and shows that the image selected by the final rerank need not be the one leading late in sampling. Under \method{} on RealLR200, the winner changes at least once on 197 of 200 images.

Figure~\ref{fig:sharing}(a) counts only the events where all four models participate (20\%, 40\%, 60\%, and 80\%, since \dpotwo{} pauses every 20\%), so the shares are directly comparable across stages. DiT4SR leads on 68\% of images at 20\%, whereas by 80\% SeeSR leads on 66\%, DiffBIR on 21.5\%, DiT4SR on 7.5\%, and \dpotwo{} on 5\%. At the 100\% final rerank, \dpotwo{} is selected on 60.5\% of images, followed by SeeSR (18\%), DiffBIR (17\%), and DiT4SR (4.5\%). The model selected at the end is therefore often not the one leading at 80\%.

Figure~\ref{fig:sharing}(b) tracks RealLR200 \#091 through three cases. Circles denote the self branch ($\mathrm{s}$, advancing without reinjection), rounded squares denote the injected branch ($\mathrm{i}$), gray nodes mark the discarded lower-scoring branch, and red outlines mark the global winner at each event. At 20\%, DiT4SR's injected branch overtakes DiffBIR, which led at 10\% (case~1). At 30\%, SeeSR wins with its self branch (case~2) and is injected into DiT4SR and DiffBIR. At 40\%, DiT4SR leads globally, while DiffBIR's injected branch outscores its self branch and is promoted to $\mathrm{s}$; this promoted branch then wins at 50\% and at the 100\% final rerank (case~3). The final output therefore descends from SeeSR's 30\% hypothesis carried forward by DiffBIR, a trajectory that final-output reranking of independent runs cannot produce.

\section{Collaborative-Pool Ablation}
\label{sec:app_pool}
Table~\ref{tab:pool} reports every pairwise and triple-model pool under the full controller (intermediate sharing, dual-path continuation, recipient-space relative transfer, and LQ-relative arbitration). Single-model results appear in Table~\ref{tab:main_results}.

\begin{table*}
\centering
\scriptsize
\setlength{\tabcolsep}{2.4pt}
\renewcommand{\arraystretch}{1.12}
\setlength{\maincol}{29.5pt}%
\caption{Collaborative-pool ablation. D: DiffBIR, S: SeeSR, T: DiT4SR, F: \dpotwo{}; \method{} denotes the full four-model system. All entries use the full controller. The best and second-best values within each row are shown in red and blue underline, respectively. Q-Align and TOPIQ-NR are held out from selection and final reranking.}
\label{tab:pool}
\begin{tabular}{@{}>{\footnotesize}w{c}{50pt}>{\footnotesize}w{c}{44pt}*{10}{>{\footnotesize}w{c}{\maincol}}>{\columncolor{oursbg}\footnotesize}w{c}{\maincol}@{}}
\toprule
\textbf{Dataset} & \textbf{Metric} & \mainhead{S+D} & \mainhead{S+T} & \mainhead{S+F} & \mainhead{D+T} & \mainhead{D+F} & \mainhead{T+F} & \mainhead{S+D+T} & \mainhead{S+D+F} & \mainhead{S+T+F} & \mainhead{D+T+F} & \mainhead{SyncShare} \\
\midrule
\multirow{6}{*}{RealLR200} & CLIPIQA+ $\uparrow$ & 0.7302 & 0.7592 & 0.7605 & 0.7512 & 0.7545 & 0.7497 & \textcolor{blue}{\underline{0.7714}} & 0.7684 & 0.7680 & 0.7649 & \textcolor{red}{\textbf{0.7738}} \\
& NIQE $\downarrow$ & 4.7397 & 4.5184 & 4.3517 & 4.8954 & 4.3365 & \textcolor{red}{\textbf{4.0724}} & 4.7998 & 4.4419 & \textcolor{blue}{\underline{4.3262}} & 4.4323 & 4.4322 \\
& MUSIQ $\uparrow$ & 70.6177 & 71.7149 & 73.3811 & 71.3865 & 73.3288 & \textcolor{red}{\textbf{73.7103}} & 71.9823 & 73.3884 & 73.5429 & \textcolor{blue}{\underline{73.6417}} & 73.5202 \\
& MANIQA $\uparrow$ & 0.5548 & 0.5827 & 0.6128 & 0.5760 & 0.6096 & 0.6041 & 0.6018 & 0.6198 & \textcolor{blue}{\underline{0.6234}} & 0.6222 & \textcolor{red}{\textbf{0.6269}} \\
& Q-Align $\uparrow$ & 4.2988 & 4.3483 & 4.4361 & 4.3306 & 4.4300 & 4.4520 & 4.3631 & 4.4487 & \textcolor{blue}{\underline{4.4608}} & \textcolor{red}{\textbf{4.4739}} & 4.4344 \\
& TOPIQ-NR $\uparrow$ & 0.6910 & 0.7006 & 0.7418 & 0.6805 & 0.7399 & 0.7410 & 0.7052 & \textcolor{red}{\textbf{0.7430}} & \textcolor{blue}{\underline{0.7429}} & 0.7383 & 0.7409 \\
\midrule
\multirow{6}{*}{RealLQ250} & CLIPIQA+ $\uparrow$ & 0.7379 & 0.7770 & 0.7887 & 0.7660 & 0.7791 & 0.7770 & 0.7848 & \textcolor{blue}{\underline{0.7906}} & 0.7905 & 0.7876 & \textcolor{red}{\textbf{0.7944}} \\
& NIQE $\downarrow$ & 4.9867 & 4.6108 & 4.2330 & 5.0420 & 4.2462 & \textcolor{red}{\textbf{3.8985}} & 4.9513 & 4.3686 & \textcolor{blue}{\underline{4.1793}} & 4.3088 & 4.3105 \\
& MUSIQ $\uparrow$ & 71.2859 & 72.8349 & 74.8565 & 72.5355 & 74.4051 & 74.7857 & 73.1504 & 74.8236 & \textcolor{blue}{\underline{74.8599}} & \textcolor{red}{\textbf{74.8610}} & 74.8550 \\
& MANIQA $\uparrow$ & 0.5280 & 0.5714 & 0.6003 & 0.5619 & 0.5917 & 0.5964 & 0.5898 & 0.6051 & 0.6086 & \textcolor{blue}{\underline{0.6101}} & \textcolor{red}{\textbf{0.6150}} \\
& Q-Align $\uparrow$ & 4.1550 & 4.3207 & 4.4740 & 4.2839 & 4.4618 & \textcolor{red}{\textbf{4.4969}} & 4.3189 & 4.4770 & 4.4894 & 4.4902 & \textcolor{blue}{\underline{4.4930}} \\
& TOPIQ-NR $\uparrow$ & 0.6829 & 0.7010 & 0.7446 & 0.6779 & 0.7372 & 0.7445 & 0.7066 & 0.7449 & \textcolor{blue}{\underline{0.7458}} & 0.7445 & \textcolor{red}{\textbf{0.7460}} \\
\midrule
\multirow{6}{*}{RealSR\_512} & CLIPIQA+ $\uparrow$ & 0.7076 & 0.7312 & 0.7250 & 0.6982 & 0.6972 & 0.6942 & \textcolor{red}{\textbf{0.7435}} & 0.7305 & 0.7280 & 0.7142 & \textcolor{blue}{\underline{0.7343}} \\
& NIQE $\downarrow$ & 4.9944 & 4.6847 & \textcolor{blue}{\underline{4.2763}} & 5.1360 & 4.6535 & \textcolor{red}{\textbf{4.1096}} & 4.9897 & 4.5485 & 4.3863 & 4.4621 & 4.5301 \\
& MUSIQ $\uparrow$ & 68.6167 & 70.1527 & 71.0232 & 68.8358 & 70.5572 & 70.6858 & 70.6708 & 71.4624 & \textcolor{blue}{\underline{71.7112}} & 71.4541 & \textcolor{red}{\textbf{71.8165}} \\
& MANIQA $\uparrow$ & 0.5482 & 0.5715 & 0.5764 & 0.5330 & 0.5681 & 0.5542 & 0.5887 & 0.5887 & \textcolor{blue}{\underline{0.5899}} & 0.5795 & \textcolor{red}{\textbf{0.6000}} \\
& Q-Align $\uparrow$ & 4.0792 & 4.1338 & \textcolor{red}{\textbf{4.1682}} & 4.0918 & 4.1145 & 4.0989 & 4.1167 & \textcolor{blue}{\underline{4.1659}} & 4.1474 & 4.1489 & 4.1560 \\
& TOPIQ-NR $\uparrow$ & 0.6816 & 0.6897 & 0.7133 & 0.6448 & 0.7050 & 0.6939 & 0.6978 & \textcolor{red}{\textbf{0.7197}} & 0.7166 & 0.7024 & \textcolor{blue}{\underline{0.7190}} \\
\midrule
\multirow{6}{*}{VideoLQ\_512} & CLIPIQA+ $\uparrow$ & 0.6195 & 0.6731 & 0.6513 & 0.6491 & 0.5862 & 0.6131 & \textcolor{red}{\textbf{0.6837}} & 0.6611 & 0.6719 & 0.6475 & \textcolor{blue}{\underline{0.6778}} \\
& NIQE $\downarrow$ & 4.5140 & 4.0669 & 4.0429 & 4.5898 & 4.7438 & \textcolor{red}{\textbf{3.8480}} & 4.4378 & 4.3196 & \textcolor{blue}{\underline{4.0071}} & 4.3221 & 4.2269 \\
& MUSIQ $\uparrow$ & 66.0383 & 68.1285 & 69.0205 & 67.5452 & 67.8417 & 69.3146 & 68.7032 & 69.3289 & \textcolor{blue}{\underline{69.8573}} & 69.5988 & \textcolor{red}{\textbf{69.9546}} \\
& MANIQA $\uparrow$ & 0.4746 & 0.5200 & 0.5154 & 0.5143 & 0.4875 & 0.4883 & \textcolor{blue}{\underline{0.5409}} & 0.5292 & 0.5351 & 0.5335 & \textcolor{red}{\textbf{0.5469}} \\
& Q-Align $\uparrow$ & 4.0771 & 4.2254 & 4.2277 & 4.1846 & 4.2015 & \textcolor{red}{\textbf{4.2765}} & 4.2073 & 4.2097 & \textcolor{blue}{\underline{4.2702}} & 4.2672 & 4.2613 \\
& TOPIQ-NR $\uparrow$ & 0.6467 & 0.6642 & 0.6846 & 0.6380 & 0.6577 & 0.6707 & 0.6701 & \textcolor{red}{\textbf{0.6886}} & 0.6879 & 0.6794 & \textcolor{blue}{\underline{0.6882}} \\
\bottomrule
\end{tabular}
\vspace{-2ex}
\end{table*}

% Placed here for page-layout control; numbered under Section C to match their discussion section.
\renewcommand{\thefigure}{C.\arabic{figure}}
\renewcommand{\thetable}{C.\arabic{table}}

\begin{figure*}
    \begin{minipage}[t]{0.48\textwidth}\centering
        \includegraphics[width=\linewidth]{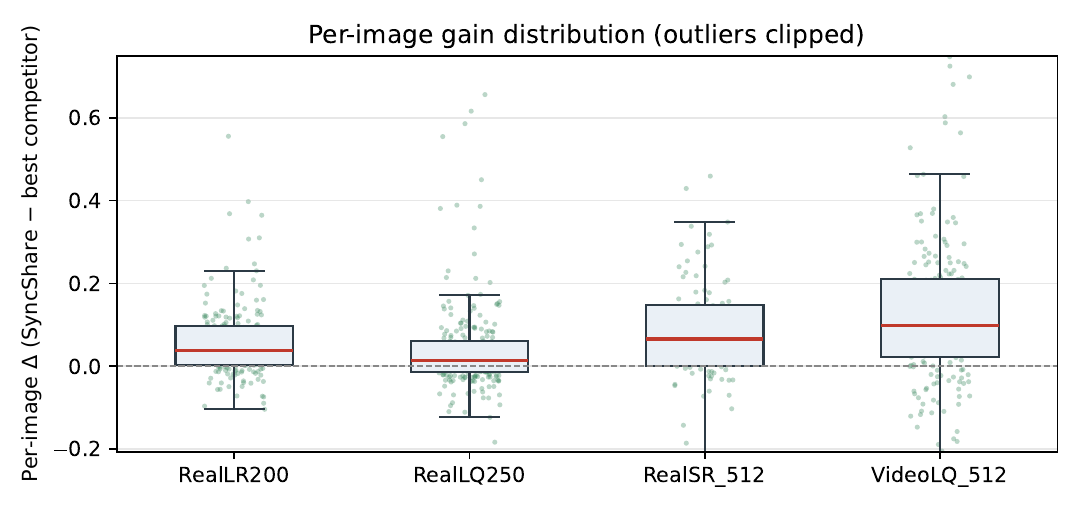}\\[2pt]
    \end{minipage}\hfill
    \begin{minipage}[t]{0.48\textwidth}\centering
        \includegraphics[width=0.715\linewidth]{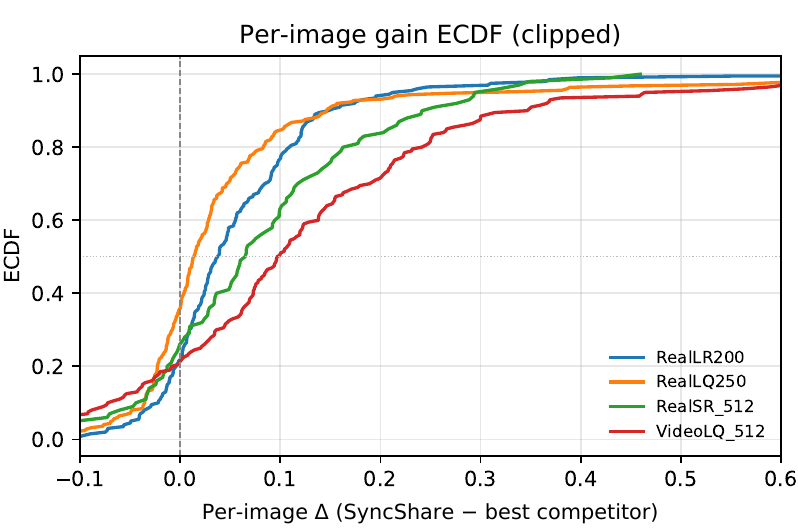}\\[2pt]
    \end{minipage}
    \caption{Per-image score difference $\Delta = S(\text{ours}) - S(\text{ref})$ between \method{} and the strongest competing method on each benchmark (\dpotwo{} on RealLR200 and RealLQ250; 4KAgent on RealSR\_512 and VideoLQ\_512). Left: box plots with individual images overlaid; the vertical axis is clipped and the dashed line marks $\Delta=0$. Right: empirical CDF of $\Delta$ (clipped to $[-0.1, 0.6]$); its value at $\Delta=0$ is the fraction of images with $\Delta \le 0$, so the win rate is one minus this value.}
    \label{fig:delta}
\end{figure*}

\begin{figure*}
  \centering
  \includegraphics[width=0.62\textwidth]{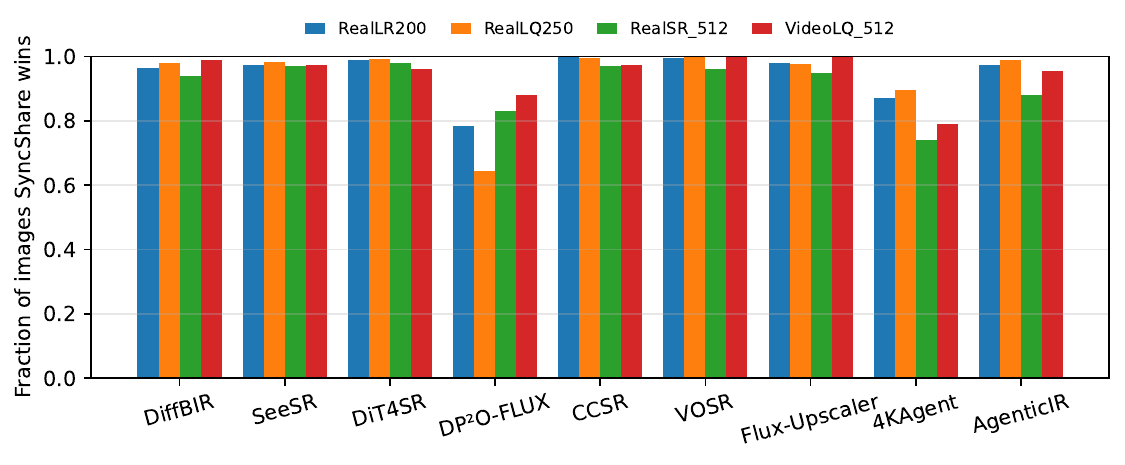}
  \vspace{-1ex}
  \caption{Fraction of images on which \method{} achieves a higher score $S$ than each competing method across the four benchmarks.}
  \label{fig:winrate}
  \vspace{-2ex}
\end{figure*}

\begin{table}
  \centering
  \scriptsize
  \setlength{\tabcolsep}{4pt}
  \renewcommand{\arraystretch}{1.15}
  \caption{Inference cost measured on six images with one NVIDIA RTX PRO 6000 (96\,GB). Time is the median wall-clock time per image, and Peak is the peak allocated GPU memory. The last row measures peak memory only for a two-model pool.}
  \label{tab:cost}
  \begin{tabularx}{\columnwidth}{c>{\centering\arraybackslash}X>{\centering\arraybackslash}Xc}
  \toprule
  \textbf{Method} & \textbf{Time (s)} & \textbf{Peak (GB)} & \textbf{Type} \\
  \midrule
  CCSR & 0.7 & 16.4 & single model \\
  DiffBIR & 20.4 & 30.4 & single model \\
  SeeSR & 14.2 & 15.3 & single model \\
  DiT4SR & 36.2 & 30.6 & single model \\
  \dpotwo{} & 11.1 & 48.3 & single model \\
  VOSR & 35.3 & 16.9 & single model \\
  Flux-Upscaler & 11.7 & 45.8 & single model \\
  AgenticIR & 37.0 & 55.5 & agent pipeline \\
  4KAgent & 224.1 & 41.9 & agent pipeline \\
  \midrule
  \rowcolor{oursbg}
  \method{} & 190.4 & 84.2 & four-model pool \\
  \midrule
  Two-model pool & -- & 57.9 & memory only \\
  \bottomrule
  \end{tabularx}
\end{table}

\begin{table*}
\centering
\scriptsize
\setlength{\tabcolsep}{5pt}
\renewcommand{\arraystretch}{1.14}
\caption{Controller controls on the strided 10-image subset of RealLR200 (\texttt{\#001}, \texttt{\#021}, \dots, \texttt{\#181}). $S$ is the composite LQ-relative score of Eq.~\eqref{eq:selector}, and $\Delta S$ is the change relative to each pool's default setting ($10\%$ normalized-progress grid, $B=2$, four-metric AnchorRank). Columns without a dagger use the four-model pool (\method{}: S+D+T+F), and those marked $\dagger$ use the three-model pool (S+T+F). The best and second-best $S$ in each column are shown in red and blue underline. The branch-budget rows repeat Table~\ref{tab:budget}.}
\label{tab:controls}
\begin{tabular}{cccccc}
\toprule
\rowcolor{headerbg}
\textbf{Control} & \textbf{Configuration} & \textbf{$S$ (4-model)}$\uparrow$ & \textbf{$\Delta S$ (4-model)} & \textbf{$S$ (3-model)}$^{\dagger}\!\uparrow$ & \textbf{$\Delta S$ (3-model)}$^{\dagger}$ \\
\midrule
\rowcolor{oursbg}
Default & $10\%$ norm.\ grid, $B=2$, four-metric AnchorRank & 0.9841 & 0.0000 & 0.9843 & 0.0000 \\
\midrule
\multirow{2}{*}{(i) Pause interval} & $20\%$ normalized grid & 0.9798 & $-0.0043$ & 0.9795 & $-0.0048$ \\
 & $40\%$ normalized grid & 0.9767 & $-0.0074$ & 0.9772 & $-0.0071$ \\
\midrule
(ii) Step alignment & Raw-step alignment (every 5 native steps) & 0.9830 & $-0.0012$ & 0.9826 & $-0.0017$ \\
\midrule
\multirow{2}{*}{(iii) Branch budget $B$} & Single branch per model ($B=1$, RelShift) & 0.9799 & $-0.0042$ & 0.9504 & $-0.0339$ \\
 & Triple branch per model ($B=3$, RelShift) & \textcolor{red}{\textbf{0.9949}} & $+0.0108$ & \textcolor{blue}{\underline{0.9913}} & $+0.0070$ \\
\midrule
\multirow{4}{*}{(iv-a) Single metric} & CLIPIQA+ only & 0.9620 & $-0.0221$ & 0.9570 & $-0.0273$ \\
 & NIQE only & 0.9015 & $-0.0826$ & 0.9021 & $-0.0822$ \\
 & MUSIQ only & 0.9685 & $-0.0156$ & 0.9766 & $-0.0077$ \\
 & MANIQA only & 0.9495 & $-0.0346$ & 0.9431 & $-0.0412$ \\
\midrule
\multirow{4}{*}{(iv-b) Leave-one-out} & w/o CLIPIQA+ (NIQE + MUSIQ + MANIQA) & \textcolor{blue}{\underline{0.9946}} & $+0.0105$ & \textcolor{red}{\textbf{0.9974}} & $+0.0131$ \\
 & w/o NIQE (CLIPIQA+ + MUSIQ + MANIQA) & 0.9785 & $-0.0056$ & 0.9582 & $-0.0261$ \\
 & w/o MUSIQ (CLIPIQA+ + NIQE + MANIQA) & 0.9750 & $-0.0092$ & 0.9791 & $-0.0052$ \\
 & w/o MANIQA (CLIPIQA+ + NIQE + MUSIQ) & 0.9680 & $-0.0161$ & 0.9539 & $-0.0304$ \\
\bottomrule
\end{tabular}
\vspace{-2ex}
\end{table*}

\renewcommand{\thefigure}{\thesection.\arabic{figure}}
\renewcommand{\thetable}{\thesection.\arabic{table}}

Comparing Table~\ref{tab:pool} against Table~\ref{tab:main_results}, every collaborative pool outperforms its best individual member on CLIPIQA+, MANIQA, and held-out TOPIQ-NR across all four benchmarks, and on MUSIQ in all cases except D+F on VideoLQ\_512 ($67.8417$ against $67.9619$ for \dpotwo{}). NIQE behaves differently: apart from S+F and T+F on RealSR\_512, no pool reaches the NIQE of its best member, reflecting the multi-metric trade-off noted in the main text. Q-Align is mixed, most noticeably on VideoLQ\_512, where seven of the eleven pools trail their best member.

Among collaborative pools, adding more models does not improve every metric. The full four-model system achieves the highest MANIQA on all four benchmarks and the highest CLIPIQA+ on RealLR200 and RealLQ250, while S+D+T leads CLIPIQA+ on RealSR\_512 and VideoLQ\_512 ($0.7435$ and $0.6837$, against $0.7343$ and $0.6778$ for \method{}). For MUSIQ, \method{} leads on RealSR\_512 and VideoLQ\_512, whereas T+F leads on RealLR200 ($73.7103$ against $73.5202$) and D+T+F on RealLQ250 ($74.8610$ against $74.8550$). T+F obtains the lowest NIQE among all pools on every benchmark, mirroring the single-model trend where DiT4SR (and VOSR on RealSR\_512) leads in NIQE (Table~\ref{tab:main_results}); NIQE thus tracks specific member models more than pool size. Q-Align likewise favors smaller pools (D+T+F on RealLR200 at $4.4739$, T+F on RealLQ250 and VideoLQ\_512 at $4.4969$ and $4.2765$, and S+F on RealSR\_512 at $4.1682$), though on RealSR\_512 and VideoLQ\_512 even these maxima stay below Flux-Upscaler ($4.2484$ and $4.3361$). Held-out TOPIQ-NR favors \method{} on RealLQ250 ($0.7460$) and S+D+F on the other three benchmarks by at most $0.0021$. Overall, on CLIPIQA+, MANIQA, TOPIQ-NR, and nearly all MUSIQ entries, the gaps among larger pools are modest compared with the gain of pooling over single restorers.

\section{Per-Image Statistics, Cost, and Controls}
\label{sec:app_stats}

\subsection{Per-Image Statistics and Cost}
For each image we compute $\Delta = S(\text{ours}) - S(\text{ref})$, where $S$ is the arbitration score of Eq.~\eqref{eq:selector} and the reference is the strongest competitor on that benchmark: \dpotwo{} on RealLR200 and RealLQ250, and 4KAgent on RealSR\_512 and VideoLQ\_512. Positive values favor \method{}. Because benchmark means can hide per-image spread, Figure~\ref{fig:delta} shows the distribution of $\Delta$ and its empirical CDF. One minus the ECDF at $\Delta=0$ yields the per-image win rates reported in the main text ($78.5\%$, $64.4\%$, $74.0\%$, and $79.0\%$). The 95\% confidence intervals of the mean difference are $[0.0475, 0.0827]$, $[0.0359, 0.0855]$, $[0.0562, 0.1050]$, and $[0.1043, 0.1549]$ on RealLR200, RealLQ250, RealSR\_512, and VideoLQ\_512. The largest per-image gains reach $1.40$, $2.30$, $0.46$, and $0.79$, and the largest losses reach $0.10$, $0.22$, $0.33$, and $0.24$; extreme values outside the clipped display ranges are not drawn. Figure~\ref{fig:winrate} breaks the win rate down by competitor. Apart from \dpotwo{} and 4KAgent, \method{} outscores every other method on at least $88\%$ of images on every benchmark (and at least $94\%$ against the other single models), reaching at least $97\%$ against CCSR and $96\%$ against VOSR.

Table~\ref{tab:cost} reports inference cost measured on six images with one NVIDIA RTX PRO 6000 (96\,GB), giving the median wall-clock time per image and the peak allocated GPU memory, with both agent pipelines run in their released configurations. Six images separate orders of magnitude rather than small timing differences. Because our implementation keeps the four restorers resident and advances them sequentially on one GPU, both runtime and memory grow with pool size. The four-model pool takes $190.4$\,s per image, close to 4KAgent ($224.1$\,s) and about $2.3\times$ the summed runtime of its four constituent models run alone ($81.9$\,s), owing to dual-branch continuation and intermediate decoding, scoring, and re-encoding at each pause. Peak memory increases from $57.9$\,GB for a two-model pool to $84.2$\,GB for four models.

\subsection{Controller Controls}
\label{sec:app_controls}
Table~\ref{tab:controls} gives the full grid behind the controller controls summarized in Section~\ref{sec:ablation}, evaluated on the uniformly strided 10-image subset of RealLR200 (\texttt{\#001}, \texttt{\#021}, \dots, \texttt{\#181}) using the score $S$ of Eq.~\eqref{eq:selector} in both the four-model pool (S+D+T+F) and the three-model pool (S+T+F). Under the pause-interval control (i), widening the interval from 10\% to 20\% and 40\% lowers $S$ monotonically from $0.9841$ to $0.9798$ and $0.9767$ with four models, and from $0.9843$ to $0.9795$ and $0.9772$ with three, so we keep the 10\% grid. Under the alignment control (ii), triggering pauses every five native steps instead of on normalized progress $\rho_k = n_k / T_k$ gives $0.9830$ and $0.9826$, trailing the progress-aligned default by $0.0012$ and $0.0017$.

Under the branch-budget control (iii), Table~\ref{tab:budget} in the main text reports the three operating points. Dropping to a single branch ($B=1$) reduces $S$ by $0.0339$ in the three-model pool compared with $0.0042$ in the four-model pool, showing that retaining the self branch ($\mathrm{s}$) alongside an injected one matters more in a smaller pool; adding a third branch ($B=3$) raises $S$ by $0.0108$ and $0.0070$, so $B=2$ is a compute--memory trade-off rather than a saturation point. Under the selector-metric control (iv), every single-metric arbiter scores well below the four-metric composite ($0.9015$--$0.9685$ with four models and $0.9021$--$0.9766$ with three), with NIQE alone lowest ($0.9015$ and $0.9021$). Removing NIQE, MUSIQ, or MANIQA also lowers $S$ in both pools ($0.9680$--$0.9785$ for four models and $0.9539$--$0.9791$ for three). Removing CLIPIQA+ instead raises $S$ on this 10-image subset ($+0.0105$ and $+0.0131$); because it contains only ten images, we do not tune the selector on it and keep all four metrics on the full benchmarks to preserve their complementary quality cues.

\section{Discussion}
\label{sec:discussion}
Sharing decoded images rather than latents allows restorers with different VAEs, schedulers, and samplers to cooperate without aligning incompatible latent spaces. Relative transfer (RelShift) encodes both the winning hypothesis and the recipient's own hypothesis with the recipient's encoder and passes only their displacement, so any shared encoder round-trip offset cancels instead of being injected as content. Across evaluation metrics (Table~\ref{tab:main_results}), NIQE favors AgenticIR or DiT4SR, whereas CLIPIQA+, MUSIQ, and MANIQA favor \method{}. Between the two held-out evaluators, TOPIQ-NR also ranks \method{} first on all four benchmarks, while Q-Align favors \method{} on RealLQ250, \dpotwo{} on RealLR200, and Flux-Upscaler on RealSR\_512 and VideoLQ\_512. Because the first four metrics also enter the selection score, the held-out metrics and visual comparisons in the main text provide the more independent evidence.

Unlike agent pipelines such as AgenticIR and 4KAgent, which invoke tools in sequence and cannot revise an earlier stage once it finishes, \method{} compares models while they are still sampling. Between consecutive pauses the active branches are independent and could execute concurrently across multiple GPUs, in which case per-event latency would be set by the slowest model plus decoding, scoring, and reinjection rather than the sum over the pool; Table~\ref{tab:cost} reflects sequential single-GPU execution without such parallelism. Finally, the controller has practical limits: it cannot recover structure when every model in the pool fails on the same region, it can propagate artifacts that no-reference metrics overrate (such as plausible but invented texture), and the four-model pool remains expensive at inference ($190.4$\,s per image and $84.2$\,GB of GPU memory).

\end{document}